%% file: main.tex
\documentclass[12pt,final]{cls/l4dc2020}

\jmlrpages{\pageref{jmlrstart}--\pageref{page:bibend}}

\usepackage[T1]{fontenc}
\usepackage{amsfonts,dsfont} 
\usepackage{microtype}
\usepackage{bm}
\usepackage[ascii]{inputenc}
\usepackage{graphicx,times,latexsym}
\usepackage{psfrag}
\usepackage{setspace}

\usepackage{color}
\usepackage{url}
\usepackage{booktabs}
\usepackage[font={small}]{caption}
\usepackage{tikz}
\newenvironment{proof}{\paragraph{Proof:}}{\hfill$\square$}
\newtheorem{defn}{Definition}
\newtheorem{Proposition}{Proposition}

\newtheorem{problem}{Problem}

\newcommand{\oprocendsymbol}{\hbox{$\bullet$}}
\newcommand{\oprocend}{\relax\ifmmode\else\unskip\hfill\fi\oprocendsymbol}

\newcommand{\real}{\mathbb{R}}

\DeclareMathOperator{\tr}{\mathbf{tr}}

\newcommand{\prl}[1]{\left(#1\right)}
\newcommand{\brl}[1]{\left[#1\right]}
\newcommand{\crl}[1]{\left\{#1\right\}}
\newcommand{\scaleMathLine}[2][1]{\resizebox{#1\linewidth}{!}{$\displaystyle{#2}$}}

\input{tex/preamble.tex}

\allowdisplaybreaks

\renewcommand{\cite}[1]{\citep{#1}}

\graphicspath{{epsfiles/}}

\usepackage{times}

\input{cls/sym.tex}

\usepackage{subfiles}  

\title[Probabilistic Safety Constraints for Learned High Relative Degree System Dynamics]{\LARGE \bf Probabilistic Safety Constraints for Learned High Relative Degree System Dynamics}

\author{%
  \Name{Mohammad Javad Khojasteh}$^{{\color{blue}*}}$ \Email{mjkhojas@caltech.edu}\\
  \addr Department of Electrical Engineering, California Institute of Technology, Pasadena, CA, 91125.
  \AND
  \Name{Vikas Dhiman}\thanks{indicates equal contribution.} \Email{vdhiman@ucsd.edu}
  \AND
  \Name{Massimo Franceschetti} \Email{mfranceschetti@ucsd.edu}
  \AND
  \Name{Nikolay Atanasov} \Email{natanasov@ucsd.edu}\\
  \addr Department of Electrical and Computer Engineering, University of California San Diego, La Jolla, CA 92093%
}

\begin{document}
\maketitle

\begin{abstract}
This paper focuses on learning a model of system dynamics online while satisfying safety constraints. Our motivation is to avoid offline system identification or hand-specified dynamics models and allow a system to safely and autonomously estimate and adapt its own model during online operation. Given streaming observations of the system state, we use Bayesian learning to obtain a distribution over the system dynamics. In turn, the distribution is used to optimize the system behavior and ensure safety with high probability, by specifying a chance constraint over a control barrier function.
\end{abstract}


\input{tex/Introduction.tex}

\input{tex/Background.tex}

\input{tex/Problem.tex}

\input{tex/InferenceNew.tex}

\input{tex/SafeControl.tex}
\input{tex/Evaluation.tex}
\input{tex/conculsion.tex}

\acks{
We gratefully acknowledge support from NSF awards CNS-1446891 and ECCS-1917177, and  support from ARL DCIST CRA W911NF-17-2-0181.
}

\def\bibliographies{bib/main}
\def\localbib{\string~/wrk/group-bib/shared}
\IfFileExists{\localbib.bib}{
  \edef\bibliographies{\bibliographies,\localbib}}{}
\IfFileExists{bib/main_filtered.bib}{
  \edef\bibliographies{\bibliographies,bib/main_filtered}}{}
\bibliography{bib/main.bib}
\label{page:bibend}

\newpage 
\appendix

\section{Notations}\label{sec:notations}
\input{./tex/note11.tex}
\section{Remarks}
\subsection{Matrix Variate Gaussians}\label{sec:MGVresutls2}
\input{./tex/MGVresults.tex}
\subsection{Further discussion about ECBF}\label{sec:ECBFresutls23}
\input{./tex/moreonECBF.tex}
\section{Proofs}
\subsection{Matrix variate Gaussian distributions}
\input{./tex/mean-and-variance-of-cbf-2/proof-mvg-cov.tex}
\input{./tex/mean-and-variance-of-cbf-2/matrix-variate-gp.tex}
\subsection{Relative degree one}\label{sec:self-triggerring}
\input{./tex/CBCerror.tex}
\input{./tex/KnonwLipshitzsefltriggering.tex}
\input{./tex/cbc-r-mean-and-variance-affine-and-quadratic.tex}
\subsection{Computing mean and variance of $\CBCtwo$ }
\input{./tex/mean-and-variance-of-cbf-2.tex}

\input{./tex/mean-and-variance-of-cbf-2/differentiating-gp.tex}
\input{./tex/mean-and-variance-of-cbf-2/differentiating-lie.tex}



%
\end{document}

%% file: tex/preamble.tex
\renewcommand{\vec}[1]{\mathbf{#1}}
\newcommand{\state}{\vec{x}}

\def\R{\mathbb{R}}
\def\E{\mathbb{E}}

\newcommand{\ctrl}{\vec{u}}

\newcommand{\stdt}{\dot{\state}}

\newcommand{\dynAff}{F}

\newcommand{\ctrlaff}{\underline{\mathbf{\ctrl}}}

\newcommand{\knl}{\kappa}

\newcommand{\StDat}{\mathbf{X}}
\newcommand{\StDtDat}{\dot{\mathbf{X}}}
\newcommand{\CtDat}{\underline{\boldsymbol{\mathcal{U}}}_{1:k}}

\newcommand{\GP}{\mathcal{GP}}

\newcommand{\grad}{\nabla}
\newcommand{\Lie}{\mathcal{L}}

\newcommand{\barf}{\bar{f}}
\newcommand{\barg}{\bar{g}}
\newcommand{\erf}{\textit{erf}}

\newcommand{\CBC}{\mbox{CBC}}
\newcommand{\CBCtwo}{\CBC^{(2)}}
\newcommand{\Prob}{\mathbb{P}}
\newcommand{\tdbff}{\bff^*_k}

\newcommand{\mDynAffs}{\bfM_k}
\newcommand{\bfBs}{\bfB_k}

\DeclareMathOperator{\vect}{\textit{vec}}
\DeclareMathOperator{\diag}{\mathbf{diag}}
\DeclareMathOperator{\cov}{cov}

\DeclareMathOperator{\Var}{\textit{Var}}

%% file: cls/sym.tex
\newcommand{\calC}{{\cal C}}
\newcommand{\calD}{{\cal D}}

\newcommand{\calG}{{\cal G}}
\newcommand{\calH}{{\cal H}}

\newcommand{\calK}{{\cal K}}

\newcommand{\calM}{{\cal M}}
\newcommand{\calN}{{\cal N}}

\newcommand{\calP}{{\cal P}}

\newcommand{\calU}{{\cal U}}

\newcommand{\calX}{{\cal X}}

\newcommand{\bfb}{\mathbf{b}}

\newcommand{\bff}{\mathbf{f}}

\newcommand{\bfk}{\mathbf{k}}

\newcommand{\bfu}{\mathbf{u}}

\newcommand{\bfx}{\mathbf{x}}
\newcommand{\bfy}{\mathbf{y}}
\newcommand{\bfz}{\mathbf{z}}

\newcommand{\bfA}{\mathbf{A}}
\newcommand{\bfB}{\mathbf{B}}
\newcommand{\bfC}{\mathbf{C}}
\newcommand{\bfD}{\mathbf{D}}

\newcommand{\bfK}{\mathbf{K}}

\newcommand{\bfM}{\mathbf{M}}

\newcommand{\bfU}{\mathbf{U}}

\newcommand{\bfX}{\mathbf{X}}
\newcommand{\bfY}{\mathbf{Y}}

\newcommand{\bfSigma}{\boldsymbol{\Sigma}}


%% file: tex/Introduction.tex
\section{Introduction}
\label{sec:intro}

Unmanned vehicles promise to transform many aspects of our lives, including transportation, agriculture, mining, and construction. 
Successful use of autonomous robots in these areas critically depends on the ability of robots to safely adapt to changing operational conditions.
Existing systems, however, rely on brittle hand-designed dynamics models and safety rules that often fail to account for both the complexity and uncertainty of real-world operation.
Recent work~\cite{deisenroth2011pilco,Dean2019,sarkar2019finite,coulson2019data,chen2018approximating,khojasteh2018learning,liu2019robust,umlauft2019feedback,fan2020deep,chowdhary2014bayesian} has demonstrated that learning-based system identification and control techniques may be successful at complex tasks and control objectives.
However, two critical considerations for applying these techniques onboard autonomous systems remain unattended: learning \textit{online}, relying on streaming data, and guaranteeing \textit{safe} operation, despite the uncertainty inherent to learning algorithms.

Motivated by the utility of Lyapunov functions for certifying stability properties, \cite{ames2016control,cbf_car,xu2015robustness,prajna2007framework,cbf} proposed \textit{Control Barrier Functions} (CBFs) as a tool for characterizing the long-term safety of dynamical systems. 
A CBF certifies whether a control policy achieves forward invariance of a \textit{safe set} $C$ by evaluating if the system trajectory remains away from the boundary of $C$. 
Most of the literature on CBFs considers systems with known dynamics, low relative degree, no disturbances, and time-triggered control, in which the control inputs are recalculated at a fixed and sufficiently small period. 
This is limiting because, low control frequency in a time-triggered setting may lead to safety constraint violation in-between sampling times.
On the other hand, high control frequency leads to inefficient use of computational resources and actuators.
\cite{yang2019self} extend the CBF framework to a self-triggered setup in which the longest time until a control input needs to be recomputed to guarantee safety is provided.
CBF techniques handle nonlinear control-affine systems but many existing results apply only to relative-degree-one systems, in which the first time derivative of the CBF depends on the control input. This requirement is violated by many underactuated robot systems and motivated extensions to relative-degree-two systems, such as bipedal and car-like robots.~\cite{hsu2015control,nguyen2016optimal}. \cite{nguyen2016exponential} generalized these ideas by designing an exponential control barrier function (ECBF) capable of handling control-affine systems with any relative degree. 

%
Providing safety guarantees for learning-based control techniques has lately been the focus of research.~\cite{learning_safe_mpc,safe_bayes_opt,fisac2018general,bastani2019safe,wabersich2018safe,biyik2019efficient}.
In particular, the CBF framework have been extend to systems with unknown dynamics.
For example, techniques for handling additive disturbances have been proposed in~\cite{clark2019control,santoyo2019barrier}, while CBF conditions for systems with uncertain dynamics have been proposed in~\cite{fan2019bayesian,wang2018safe,taylor2019adaptive,cheng2019end,salehi2019active}.
Furthermore, \cite{fan2019bayesian} study time-triggered CBF-based controllers for control-affine systems with relative degree one, where the input gain part of the dynamics is known and invertible.
Bayesian learning is used in~\cite{fan2019bayesian}
to determine a distribution over the drift term of the dynamics.
In particular,~\cite{fan2019bayesian} compared the performances of Gaussian Process regression~\cite{williams2006gaussian}, Dropout neural networks~\cite{gal2016dropout}, and ALPaCA~\cite{harrison2018meta} in simulations.
\cite{wang2018safe},~\cite{cheng2019end}, and~\cite{taylor2019adaptive} have studied time-triggered CBF-based control relative-degree-one systems in presence of additive uncertainty in the drift part of the dynamics.
In~\cite{wang2018safe}, GP regression is used to approximate the unknown part of the 3D nonlinear dynamics of a quadrotor.
\cite{cheng2019end} proposed a two-layers control design architecture that integrates CBF-based controllers with model-free reinforcement learning.
\cite{taylor2019adaptive} proposed adaptive CBFs to deal with parameter uncertainty. \cite{salehi2019active} studies nonlinear systems only with drift terms and uses Extreme Learning Machines to approximate the dynamics.

Our work proposes a learning approach for estimating posterior distribution of robot dynamics from online data to design a control policy that guarantees safe operation. We make the following \textbf{contributions}. First, we develop a matrix variate Gaussian Process (GP) regression approach with efficient covariance factorization to learn the \textit{drift term} and \textit{input gain} terms of a nonlinear control-affine system. Second, we use the GP posterior to specify a probabilistic safety constraint and determine the longest time until a control input needs to be recomputed to guarantee safety with high probability. Finally, we extend our formulation  
to dynamical systems with arbitrary relative degree and show that a safety constraint can be specified only in terms of the mean and variance of the Lie derivatives of the CBF. \textbf{Notation, proofs, and additional remarks are available in the appendix at arXiv~\cite{usfinal23}}.

%% file: tex/Background.tex
\section{Background}
\label{sec:background}

Consider a control-affine nonlinear system:
\begin{align}
\label{eq:system_dyanmics}
\dot{\bfx} = f(\bfx) + g(\bfx)\bfu = \begin{bmatrix} f(\bfx) & g(\bfx)\end{bmatrix} \begin{bmatrix}1\\\bfu\end{bmatrix} =: F(\bfx) \ctrlaff
\end{align}
where $\bfx(t) \in \real^n$ and $\bfu(t)\in \real^m$ are the system state and control input, respectively, at time $t$. Assume that the \textit{drift term} $f: \real^n \rightarrow \real^n$ and the \textit{input gain} $g: \real^n \rightarrow \real^{n \times m}$ are locally Lipschitz. We study the problem of enforcing probabilistic safety properties via CBF when $f$ and $g$ are unknown. We first review key results on CBF-based safety for \textit{known dynamics}~\cite{cbf}.

\subsection{Known Dynamics: Control Barrier Functions for Safety}
\label{sec:cbf}
Let $\mathcal{C} \subset \mathcal{D} \subset \mathbb{R}^n$ be a \textit{safe set} of system states. Assume $\calC = \{\bfx \in \mathcal{D} \mid h(\bfx) \geq 0\}$ is specified as the superlevel set of $h \in \calC^1(\calD,\real)$, a continuously differentiable function $\mathcal{D} \rightarrow \mathbb{R}$, such that $\grad_\bfx h(\bfx) \neq 0$ for all $\bfx$ when $h(\bfx)=0$. For any initial condition $\bfx(0)$, there exists a maximum time interval $I(\bfx(0))=[0,\bar{t})$ with $\bar{t} \in \mathbb{R} \cup \{\infty\}$ such that $\bfx(t)$ is a unique solution to~\eqref{eq:system_dyanmics}~\cite{Khalil:1173048}. System~\eqref{eq:system_dyanmics} is \textit{safe} with respect to set $\calC$ if $\calC$ is \textit{forward invariant}, i.e., for any $\bfx(0) \in \calC$, $\bfx(t)$ remains in $\calC$ for all $t$ in $I(\bfx(0))$. System safety may be asserted as follows.


\begin{defn}
\label{def:cbf}
A function $h \in \calC^1(\calD,\real)$ is a control barrier function (CBF) for the system in~\eqref{eq:system_dyanmics} if the control barrier condition (CBC), 
$\sup_{\bfu}
\mbox{CBC}(\bfx, \bfu) \geq 0 $, is satisfied for all $\bfx \in \mathcal{D}$;
where $\mbox{CBC}(\bfx, \bfu) := \Lie_{f}h(\bfx) + \Lie_{g}h(\bfx)\bfu + \alpha(h(\bfx))$, $\alpha$ is any extended class $K_\infty$ function and $\Lie_fh(\bfx)$ and $\Lie_gh(\bfx)$ are the Lie derivatives of $h$ along $f$ and $g$, respectively.
\end{defn}


\begin{theorem}[Sufficient Condition for Safety~\cite{cbf}]
\label{thm:Ames1}
Consider a safe set $\calC$ with associated function $h \in \calC^1(\calD,\real)$. If $\grad_\bfx h(\bfx) \neq 0$ for all $\bfx \in \partial \mathcal{C}$, then any Lipschitz continuous control policy $\pi(\bfx) \in \crl{\bfu \in \calU \mid \mbox{CBC}(\bfx,\bfu) \geq 0}$ renders the system in~\eqref{eq:system_dyanmics} safe.
\end{theorem}
\citet{cbf} also provide a necessary condition for safety allowing a concise charaterization:
\begin{align}
\text{~\eqref{eq:system_dyanmics} is safe with respect to}~\mathcal{C}~~\Leftrightarrow~~\exists~\bfu = \pi(\bfx)~~\text{s.t.}~~\mbox{CBC}(\bfx,\bfu)\ge 0 \;~ \forall \bfx \in \mathcal{D}.
\end{align}


\subsection{Known Dynamics: Optimization-based Safe Control}
The results in Sec.~\ref{sec:cbf} allow designing a control policy $\pi(\bfx)$ that guarantees system safety as long as $\mbox{CBC}(\bfx,\pi(\bfx))$ remains positive at all times. In practice, this is achieved by solving a quadratic program (QP) repeatedly at triggering times $t_k = k \tau$ for $k \in \mathbb{N}$ and $\tau > 0$:
\begin{align}
\label{opt-cbf}
\min_{\bfu_k}
\;\;& \bfu_k^\top Q\bfu_k
\qquad\text{s.t.}~~\mbox{CBC}(\bfx_k,\bfu_k) \ge 0,
\end{align}
where $Q \succ 0$, $\bfx_k := \bfx(t_k)$, $\bfu_k := \bfu(t_k)$. While the QP above cannot be solved infinitely fast, Theorem $3$ of~\citet{ames2016control} shows that if $f$, $g$, and $\alpha \circ h$ are locally Lipschitz, then $\bfu_k(\bfx)$ and $\mbox{CBC}(\bfx,\bfu_k(\bfx))$ are locally Lipschitz. Thus, for sufficiently small $\tau$, solving~\eqref{opt-cbf} at $\{t_k\}_{k \in \mathbb{N}}$ ensures safety during the inter-triggering times as well.

%% file: tex/Problem.tex
\section{Problem Statement}
\label{sec:problem}


Consider a control-affine nonlinear system~\eqref{eq:system_dyanmics}, where $F: \real^n \rightarrow \real^{n \times (m+1)}$ is \textit{unknown}. Our objective is to estimate $F(\bfx)$ from online observations of the system state and control trajectory and ensure that~\eqref{eq:system_dyanmics} remains safe with respect to a set $\calC$.

\begin{problem}
\label{prb:learning}
Given a prior Gaussian Process distribution $\vect(F(\bfx))$\footnote{$\vect(F(\bfx)) \in \real^{n(m+1)}$ is a vector obtained by stacking the columns of $F(\bfx)$}$\sim \calG\calP\prl{\vect(\bfM_0(\bfx)), \bfK_0(\bfx,\bfx')}$ on the unknown system dynamics and a training set $\StDat_{1:k} := [\bfx(t_1), \dots, \bfx(t_k)]$, $\bfU_{1:k} := [\bfu(t_1),\allowbreak \dots, \bfu(t_k)]$, $\StDtDat_{1:k}=[\dot{\bfx}(t_1), \dots, \dot{\bfx}(t_k)]$\footnote{If not available, the derivatives may be approximated via $\StDtDat_{1:k-1} := \bigl[
  \frac{\bfx(t_2) - \bfx(t_1)}{t_2-t_1}^\top, \dots, \frac{\bfx(t_k) -
    \bfx(t_{k-1})}{t_{k} - t_{k-1}}^\top \bigr]^\top$ provided that the inter-triggering times $\{\tau_k\}$ are sufficiently small.}, compute the posterior Gaussian Process distribution $\calG\calP\prl{\vect(\bfM_k(\bfx)), \bfK_k(\bfx,\bfx')}$ of $\vect(F(\bfx))$ conditioned on $(\StDat_{1:k}, \bfU_{1:k}, \StDtDat_{1:k})$.
\end{problem}

\begin{problem}
\label{prob3444!!}
Given a safe set $\calC$, and a safe system state $\bfx_k := \bfx(t_k) \in \calC$, and the distribution $\calG\calP(\textit{vec}(\bfM_k(\bfx)),\allowbreak \bfK_k(\bfx,\bfx'))$ of $\vect(F(\bfx))$ at time $t_k$, choose a control input $\bfu_k$ and triggering period $\tau_k$ such that:
\begin{equation}
\label{eorprobmelm45}
\mathbb{P}(\mbox{CBC}(\bfx(t),\bfu_k) \ge 0) \ge p_k \quad \text{for} \quad \bfu(t) \equiv \bfu_k \quad\text{and}\quad t \in [t_k,t_k+\tau_k)
\end{equation}
where $\bfx(t)$ follows the dynamics in~\eqref{eq:system_dyanmics}, and $p_k \in (0,1)$ is a user-specified risk tolerance.
\end{problem}

%% file: tex/InferenceNew.tex
\section{Matrix Variate Gaussian Process Regression of System Dynamics}
\label{learningsec:13452}

We propose an efficient
Gaussian Process (GP) regression approach to estimate a posterior distribution over the dynamics $F(\bfx)$ of the nonlinear control-affine systems~\eqref{eq:system_dyanmics}. The posterior will be used to determine the distribution of $\mbox{CBC}(\bfx,\bfu)$ in Sec.~\ref{sec:unknowndynamics}\footnote{We only consider epistemic but no aleatoric uncertainty. Namely, while $F(\bfx)$ is sampled from a GP, no additive disturbances are considered for the dynamics~\eqref{eq:system_dyanmics}.}. 
Since $F(\bfx)$ is matrix-valued, we define a GP over its columnwise vectorization, $\vect(F(\bfx)) \sim \mathcal{GP}(\vect(\bfM_0(\bfx)), \bfK_0(\bfx,\bfx'))$. The controller can observe $\StDat_{1:k}$ and $\bfU_{1:k}$ without noise, but the measurements $\StDtDat_{1:k}$ might be noisy. As the controller observes $f(\bfx)$ and $g(\bfx)$ together via $\StDtDat_{1:k}$, there may be a correlation between their different components. 
Thus, we develop an efficient factorization of $\bfK_0(\bfx,\bfx')$ based on the Matrix Variate Gaussian distribution~\cite{StructuredPBP,louizos2016structured} to learn $f(\bfx)$ and $g(\bfx)$ together.
We provide definition and properties of the MVG distribution in Appendix~\ref{sec:MGVresutls2}. Two alternative approaches to infer a posterior over $F(\bfx)$ and their drawbacks are also discussed in Appendix~\ref{sec:MGVresutls2}.

Note that if $\bfX \sim \calM\calN(\bfM,\bfA,\bfB)$, then $\vect(\bfX) \sim \mathcal{N}(\vect(\bfM), \bfB \otimes \bfA)$. Based on this observation, we propose the following GP parameterization for the vector-valued functions $\vect(F(\bfx))$:
\begin{equation}
\begin{aligned}
\vect(F(\bfx)) &\sim \mathcal{GP}(\vect(\bfM_0(\bfx)), \bfB_0(\bfx,\bfx') \otimes \bfA)\\
\end{aligned}
\end{equation}
The above parameterization is efficient as compared to learning the full covariance $\bfK_0(.,.) \in \R^{n(m+1)\times (m+1)n}$, because we need to learn smaller matrices, $\bfB_0(\bfx,\bfx') \in \mathbb{R}^{(m+1)\times (m+1)}$ and $\bfA \in \mathbb{R}^{n \times n}$.
 Fortunately, this parameterization also preserves its structure on inference.

Consider the training set $(\StDat_{1:k}, \bfU_{1:k}, \StDtDat_{1:k})$ and a query test point $\bfx_*$. The train and test data are jointly Gaussian:
\begin{align*}
  \scaleMathLine{\begin{bmatrix}
  \dot{\bfx}_1\\\vdots\\\dot{\bfx}_k\\\vect(F(\bfx_*))
  \end{bmatrix} \sim \mathcal{N}\prl{ \begin{bmatrix} \bfM_0(\bfx_1)\underline{\bfu}_1\\\vdots\\\bfM_0(\bfx_k)\underline{\bfu}_k\\\vect(\bfM_0(\bfx_*))\end{bmatrix}, \begin{bmatrix} \underline{\bfu}_1^\top \bfB_0(\bfx_1,\bfx_1) \underline{\bfu}_1& \cdots & \underline{\bfu}_1^\top \bfB_0(\bfx_1,\bfx_k) \underline{\bfu}_k & \underline{\bfu}_1^\top\bfB_0(\bfx_1,\bfx_*) \\ \vdots&\ddots&\vdots&\vdots\\ \underline{\bfu}_k^\top \bfB_0(\bfx_k,\bfx_1) \underline{\bfu}_1& \cdots & \underline{\bfu}_k^\top \bfB_0(\bfx_k,\bfx_k) \underline{\bfu}_k & \underline{\bfu}_k^\top\bfB_0(\bfx_k,\bfx_*)\\\bfB_0(\bfx_*,\bfx_1) \underline{\bfu}_1& \cdots & \bfB_0(\bfx_*,\bfx_k) \underline{\bfu}_k & \bfB_0(\bfx_*,\bfx_*)\end{bmatrix} \otimes \bfA}}.
\end{align*}
\sloppy 
In the above formulation, the resulting posterior is independent of query control input, $\bfu_*$, which allows us to use this posterior in Sec.~\ref{sec:unknowndynamics} to efficiently compute a safe control input.
To simplify notation, let $\bfB_0(\bfX_{1:k},\bfX_{1:k}) \in \real^{k(m+1) \times k(m+1)}$ be a matrix with elements $\brl{\bfB_0(\bfX_{1:k},\bfX_{1:k})}_{ij} := \bfB_0(\bfx_i,\bfx_j)$ and define $\boldsymbol{\mathcal{M}}_{1:k}:= \begin{bmatrix} \bfM_0(\bfx_1) & \cdots & \bfM_0(\bfx_k) \end{bmatrix} \in \real^{n \times k(m+1)}$ and $\underline{\boldsymbol{\mathcal{U}}}_{1:k}:= \diag(\ctrlaff_1, \dots, \ctrlaff_k) \in \R^{k(m+1) \times k}$. Applying a Schur complement, we can derive the posterior distribution of $\vect(F(\bfx_*))$ conditioned on $(\StDat_{1:k}, \bfU_{1:k}, \StDtDat_{1:k})$ as a Gaussian Process $\mathcal{GP}(\vect(\bfM_k(\bfx_*)), \bfB_k(\bfx_*,\bfx_*')\otimes\bfA)$ with parameters:
\begin{equation*}
\begin{aligned}
\bfM_k(\bfx_*) &:= \bfM_0(\bfx_*) + \prl{ \dot{\bfX}_{1:k} - \boldsymbol{\mathcal{M}}_{1:k}\underline{\boldsymbol{\mathcal{U}}}_{1:k}} \prl{\underline{\boldsymbol{\mathcal{U}}}_{1:k}^\top\bfB_0(\bfX_{1:k},\bfX_{1:k})\underline{\boldsymbol{\mathcal{U}}}_{1:k}}^{-1}\underline{\boldsymbol{\mathcal{U}}}_{1:k}^\top\bfB_0(\bfX_{1:k},\bfx_*)\\
\bfB_k(\bfx_*,\bfx_*') &:= \bfB_0(\bfx_*,\bfx_*') + \bfB_0(\bfx_*,\bfX_{1:k})\underline{\boldsymbol{\mathcal{U}}}_{1:k}\prl{\underline{\boldsymbol{\mathcal{U}}}_{1:k}^\top\bfB_0(\bfX_{1:k},\bfX_{1:k})\underline{\boldsymbol{\mathcal{U}}}_{1:k}}^{-1}\underline{\boldsymbol{\mathcal{U}}}_{1:k}^\top\bfB_0(\bfX_{1:k},\bfx_*')
\label{eq:mvg-posterior}
\end{aligned}
\end{equation*}
This inference has a computation complexity of $O((1 +m)^3k^2) +O(k^3)$ 
while the same for independent GP is $O((1 +m)k^2) +O(k^3)$. Since $k >> m$ is common, the proposed model has almost same inference cost as independent GP.
Step by step details are provided in Appendix~\ref{proof:matrix-variate-gp}. For a given query control input $\bfu_*$, the posterior of $F(\bfx_*)\underline{\bfu}_*$ is:
\begin{equation}
\label{eq:gp_posterior}
F(\bfx_*)\underline{\bfu}_* = f(\bfx_*) + g(\bfx_*)\bfu_* \sim \mathcal{GP}(\bfM_k(\bfx_*)\underline{\bfu}_*, \underline{\bfu}_*^\top\bfB_k(\bfx_*,\bfx_*')\underline{\bfu}_*\otimes\bfA).
\end{equation}

%% file: tex/SafeControl.tex
\section{Self-triggered Control with Probabilistic Safety Constraints}
\label{sec:unknowndynamics}
Sec.~\ref{learningsec:13452} addressed Problem~\ref{prb:learning} by proposing an efficient Gaussian Process inference algorithm for nonlinear control-affine systems. Now, we consider Problem~\eqref{prob3444!!}. As discussed in Sec.~\ref{sec:cbf} if $f$ and $g$ are locally Lipschitz, then system~\eqref{eq:system_dyanmics} has a unique solution for any $\bfx(0)$ for all time $t$ in $I(\bfx(0))$. We assume the sample paths of the GP used to model the dynamics~\eqref{eq:system_dyanmics} are locally Lipschitz with high probability. Similar smoothness assumption has been made previously in~\citet{srinivas2009gaussian}. As mentioned in Problem~\eqref{prob3444!!}, we use a zero-order hold (ZOH) control mechanism in inter-triggering time, i.e., $\bfu(t) \equiv \bfu_k$ for $t \in [t_k,t_k+\tau_k)$. In detail, we assume that for any $L_k >0$, $\bfu_k$, and triggering time $t_k$, there exists a constant $b_k >0$, such that,
\begin{align}
\label{eq:smoth23}
\Prob\left(
\sup_{s \in [0, \tau_k)}\|F(\bfx(t_k+s))\ctrlaff_k
   -F(\bfx_k)\ctrlaff_k\| \le L_k \|\bfx(t_k+s)-\bfx_k\|
   \right) \ge q_k:=1-e^{-b_kL_k}.
\end{align}
This assumption is valid for a large class of GPs, e.g., those with stationary kernels that are four times differentiable, such as squared exponential and some Mat{\'e}rn kernels~\cite{ghosal2006posterior,shekhar2018gaussian}. However, it may not hold for GPs with highly erratic sample paths. 

The posterior of $F(\bfx)\bfu$ in~\eqref{eq:gp_posterior} induces a distribution over $\mbox{CBC}(\bfx,\bfu)$. To ensure that safety in the sense of~\eqref{eorprobmelm45} is preserved over a period of time $[t_k,t_k+\tau_k)$, we enforce a tighter constraint at time $t_k$ and determine the time $\tau_k$ for which it remains valid. In detail, we solve a chance-constrained version of~\eqref{opt-cbf} at time $t_k$,
\begin{equation}
\label{prog:CBF-CLF:unknown}
\min_{\bfu_k} \bfu_k^\top Q\bfu_k \qquad \text{s.t.}~~\mathbb{P}(\mbox{CBC}(\bfx_k,\bfu_k) \ge \zeta | \bfx_k,\bfu_k) \ge \tilde{p}_k,
\end{equation}
where $\tilde{p}_k = p_k/q_k$. The choice of $\zeta$ and its effect on $\tau_k$ is discussed next.

\begin{lemma}
\label{lem:NRcalculation}
Consider the dynamics in~\eqref{eq:system_dyanmics} with posterior distribution in~\eqref{eq:gp_posterior}. Given $\bfx_k$ and $\bfu_k$, $\mbox{CBC}_k := \mbox{CBC}(\bfx_k,\bfu_k)$ is a Gaussian random variable with the following parameters:
\begin{align}
\label{eq:parametofpi5543}
\E[\mbox{CBC}_k] &= \nabla_\bfx h(\bfx_k)^\top \bfM_k(\bfx_k)\underline{\bfu}_k + \alpha(h(\bfx_k)),\\
\Var[\mbox{CBC}_k] &=  \underline{\bfu}_k^\top\bfB_k(\bfx_k,\bfx_k)\underline{\bfu}_k \nabla_\bfx h(\bfx_k)^{\top}\bfA\nabla_\bfx h(\bfx_k)
\end{align}
\end{lemma}
Using Lemma~\ref{lem:NRcalculation}, we can rewrite the safety constraint as
\begin{equation}
\mathbb{P}(\mbox{CBC}_k \ge \zeta | \bfx_k,\bfu_k) = 1 - \Phi\prl{\frac{\zeta-\E[\mbox{CBC}_k]}{\sqrt{\Var[\mbox{CBC}_k]}}} \geq \tilde{p}_k, 
\end{equation}
where $\Phi(\cdot)$ is the cumulative distribution function of the standard Gaussian.
Note that if the control input is chosen so that $\zeta-\E[\mbox{CBC}_k]<0$, as the posterior variance of $\mbox{CBC}_k$ tends to zero, the probability $\mathbb{P}(\mbox{CBC}_k \ge \zeta | \bfx_k,\bfu_k)$ tends to one.
Namely, as the uncertainty about the system dynamics tends to zero, our results reduce to the setting of Sec.~\ref{sec:cbf}, and safety can be ensured with probability one.
Noting that $\Phi^{-1}(1-\tilde{p}_k) = \sqrt{2}\mbox{erf}^{-1}(1-2\tilde{p}_k)$,
controller \eqref{prog:CBF-CLF:unknown} can be rewritten as
\begin{align}\label{lower:nu1}
\min_{\bfu_k} \bfu_k^\top Q\bfu_k
~~\text{ s.t. }~~
\E[\mbox{CBC}_k] - \zeta \ge 0
\text{ and }
(\E[\mbox{CBC}_k] - \zeta)^2 \ge 2\Var[\mbox{CBC}_k]~(\mbox{erf}^{-1}(1-2\tilde{p}_k))^2.
\end{align}

The program~\eqref{lower:nu1} provides a probabilistic safety constraints at the triggering times $\{t_k\}_{k \in \mathbb{N}}$. Next, we will extend our analysis to inter-triggering times $\{\tau_k\}$.
We continue by re-writing the Proposition $1$ of~\cite{yang2019self} for our setup.
\begin{Proposition}
\label{self-trige3Lip}
Consider the system in~\eqref{eq:system_dyanmics} with zero-order hold control in inter-triggering times. If the event~\eqref{eq:smoth23} occurs at the $k$th triggering time, then  for all  $s \in [0, \tau_k)$ we have
\begin{align}\label{defnirrrt5654}
   &\|\bfx(t_k+s)-\bfx_k\| \le  \overline{r}_k(s):=\frac{1}{L_k}\|\dot{\bfx}_k\|\left(e^{L_ks}-1\right).
\end{align}
\end{Proposition}
Recall from Sec.~\ref{sec:cbf} that $h$ is  a continuously differentiable function. Thus using Proposition~\ref{self-trige3Lip}, we notice for any inter-triggering time $\tau_k$, there exist a constant $\chi_k > 0$ such that  
\begin{align}
\label{nirn3453456543!}
    \sup_{s \in [0, \tau_k)}\|\nabla h(\bfx(t_k+s))\| \le \chi_k.
\end{align}
This is used in the next theorem which concerns Problem~\ref{prob3444!!}.

\begin{theorem}
\label{ir5o45t!!!}
Consider the system in~\eqref{eq:system_dyanmics} with safe set $\mathcal{C}$.
Assume the program~\eqref{prog:CBF-CLF:unknown} has a solution at triggering time $t_k$, event~\eqref{eq:smoth23} occurs at least with probability $q_k$, $\|\dot{\bfx}_k\|\neq0$, and for all $s \in [0, \tau_k)$, $\alpha \circ h$ satisfies the following Lipschitz property
\begin{align}
\label{htym6!7uytf}
    |\alpha \circ h(\bfx(t_k+s))-\alpha \circ h(\bfx_k)| 
    \le L_{\alpha \circ h} \|\bfx(t_k+s)-\bfx_k\|.
\end{align} 
Then~\eqref{eorprobmelm45} is valid for $p_k=\tilde{p}_kq_k$, and $\tau_k \le \frac{1}{L_k}\ln\left(1+\frac{L_k\zeta}{(\chi_kL_k+L_{\alpha \circ h})\|\dot{\bfx}_k\|}\right)$, where $\chi_k$ is given in~\eqref{nirn3453456543!}.
\end{theorem}

\begin{remark}
\label{remar1!}
\rm{
Assuming  $\|\dot{x}(t_k)\|\neq0$ in Theorem~\eqref{ir5o45t!!!} is not restricting our results. Since, if the state of the system is safe and it does not change it remains safe. 
 }
 \end{remark}


\section{Extension to Higher Relative-degree Systems}
\label{sec:exp-cbf-rel-deg-2}
\newcommand{\tdLie}{\hat{\Lie}}
\newcommand{\bLie}{\bar{\Lie}}
\newcommand{\bfxdot}{\dot{\bfx}}
\newcommand{\ff}{\mathfrak{\zeta}}


Next, we extend the probabilistic safety constraint formulation for systems with arbitrary relative degree, using an exponential control barrier function (ECBF)~\cite{nguyen2016exponential,cbf}~\footnote{The motivation for assuming known relative degree and CBF but unknown dynamics comes from robotics applications. Commonly, the class of the system is known
but the parameters (e.g., mass, the moment of inertia) and high-order interactions (e.g., jerk, snap) of the dynamics are unknown.  Finding the relative degree and a proper CBF is left open for future work (cf.~\cite{akella2020formal,robey2020learning}).} 

Let $r \ge 1$ be the relative degree of $h(\bfx)$, that is, $\Lie_g\Lie_f^{(r-1)}h(\bfx) \ne 0$ and $\Lie_g\Lie_f^{(k-1)}h(\bfx) = 0$, $\forall k \in \{1, \dots, r-2\}$.
  Define traverse dynamics with traverse vector $\eta(\bfx)$,
  \begin{align}
    \label{eq:traverse-system}
    & \dot{\eta}(\bfx) = \mathcal{F}\eta(\bfx) + \mathcal{G}\bfu,
    & h(\bfx) = C\eta(\bfx)
  \end{align}
  where $C = [1, 0, \dots, 0]^\top \in \R^r$. Also, $ \eta(\bfx)$, $\mathcal{F}$, and $\mathcal{G}$ are defined in Appendix~\ref{sec:notations}.
\begin{defn}
\label{def:cbf-high}
A function $h \in \calC^r(\calD,\real)$ is an exponential control barrier function (ECBF) for the system in~\eqref{eq:system_dyanmics} if there exists a row vector $K_\alpha \in \R^r$ such that the $r$th order condition $\mbox{CBC}^{(r)}(\bfx, \bfu) := \Lie_f^{(r)}h(\bfx) + \Lie_g\Lie_f^{(r-1)}h(\bfx)\bfu+K_\alpha \eta(\bfx)$ satisfies 
$ \sup_{\bfu} \mbox{CBC}^{(r)}(\bfx, \bfu) \geq 0$ for all $ \bfx \in \mathcal{D}$,
which results in $h(\bfx(t)) \ge C\eta(\bfx_0)e^{(\mathcal{F}-\mathcal{G}K_\alpha)t}  \ge 0$, whenever $h(\bfx_0) \ge 0$.
\end{defn}
If $K_\alpha$ is chosen appropriately (see Appendix~\ref{sec:ECBFresutls23}), a control policy $\bfu = \pi(\bfx)$ that ensures $\mbox{CBC}^{(r)} \ge 0$, renders the dynamics~\eqref{eq:system_dyanmics} safe with respect to set $\mathcal{C}$. 
%
  %
  %
Thus, as in~\eqref{prog:CBF-CLF:unknown}, we are interested in solving
\begin{align}
\label{progg3456755!!}
\min_{\bfu_k} \;\; \bfu_k^\top Q\bfu_k \qquad
\text{s.t.}~~ &\mathbb{P}(\mbox{CBC}_k^{(r)} \ge \zeta | \bfx_k,\bfu_k) \ge \tilde{p}_k.
\end{align}

\begin{Proposition}
\label{prob:VarUquadrac24562}
For a control-affine system of relative degree $r$, the expectation $\E[\mbox{CBC}_k^{(r)}]$ is affine in $\bfu$ and $\Var[\mbox{CBC}_k^{(r)}]$ is quadratic in $\bfu$ (Proof in~Sec~\ref{proof:cbc-r-mean-affine-var-quadratic}).
\end{Proposition}

\begin{Proposition}
\label{prop:456723o433302}
For a control-affine system of relative degree $r$, as defined in~\eqref{eq:system_dyanmics}, the system stays in the safe set $\calC$ with ECBF $h$ if the control is determined from the following Quadratically Constrained Quadratic Program (QCQP) (Proof in Sec~\ref{proof:cbc-r-quad-prog}),
\begin{align}
\label{progg!!5544i}
\min_{\bfu_k
} \;\;& \bfu_k^\top Q\bfu_k
&
\text{s.t.}~~~~ &\E[\mbox{CBC}_k^{(r)}] - \zeta \ge 0~~~~
\text{and}~~
(\E[\mbox{CBC}_k^{(r)}] -\zeta)^2 \ge \frac{\tilde{p}_k}{1-\tilde{p}_k} \Var[\mbox{CBC}_k^{(r)}]
\end{align}
\end{Proposition}

Solving the program~\eqref{progg!!5544i} requires the knowledge of the mean and variance of $\mbox{CBC}_k^{(r)}$
(see Thm.~\ref{thm:CBC2-quanti34} in Appendix~\ref{proof:mean-var-of-cbc} for $\CBCtwo$).
In general, Monte Carlo sampling could be used to estimate these quantities.
The chance constraint in~\eqref{progg!!5544i} 
can be interpreted the standard deviation of $\mbox{CBC}_k^{(r)}$ should be smaller than the mean by a factor of $\sqrt{\tilde{p}_k/(1-\tilde{p}_k)}$.

%% file: tex/Evaluation.tex
\section{Simulations}
\begin{figure}
\centering
\includegraphics[width=0.22\linewidth]{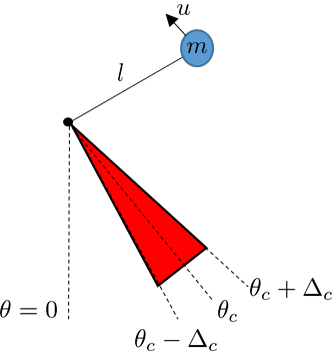}%
\includegraphics[width=0.66\linewidth]{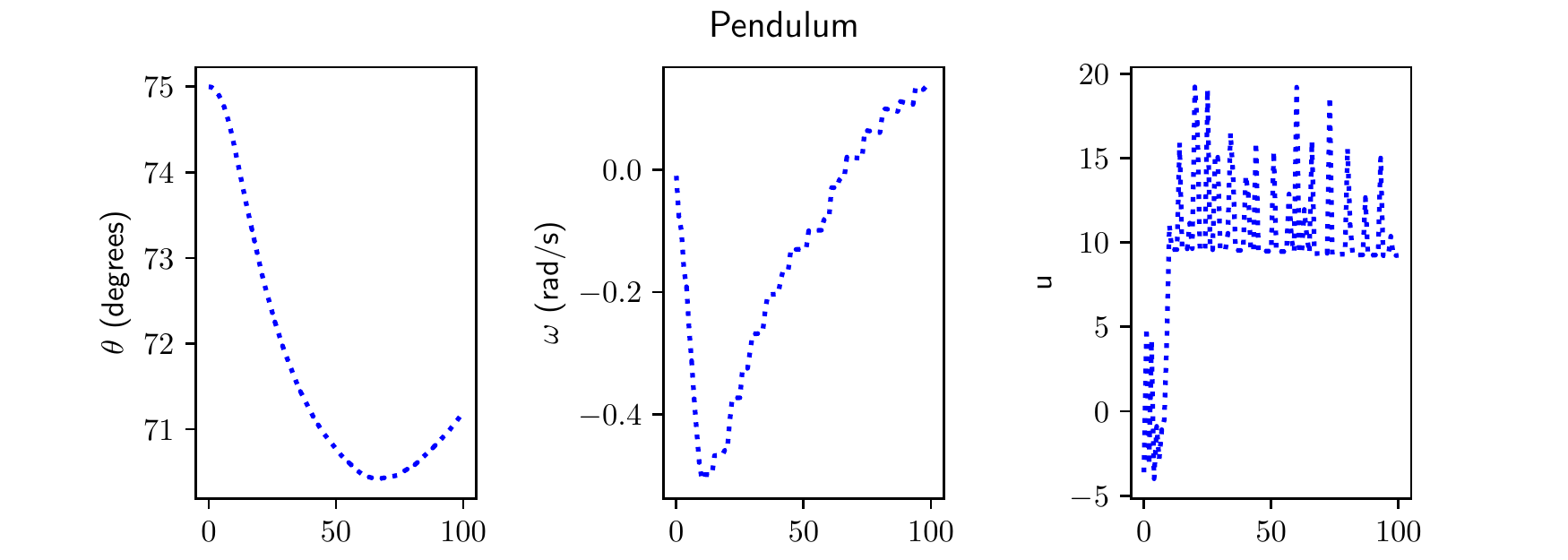}\\
    \includegraphics[width=\linewidth]{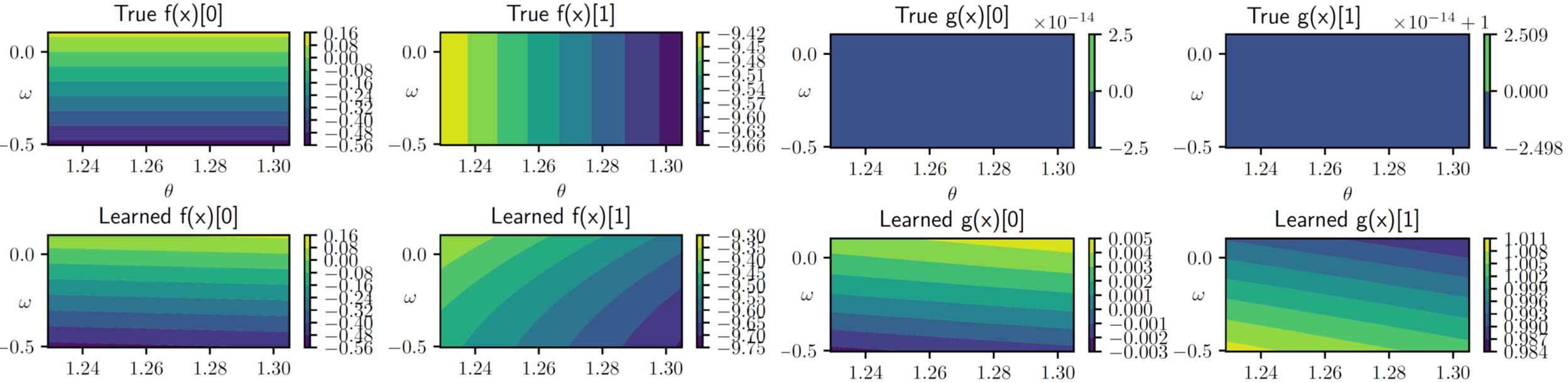}
\caption{\textbf{Top left}: Pendulum simulation (left) with an unsafe (red) region. \textbf{Top right}: The pendulum trajectory (middle) resulting from the application of safe control inputs (right) is shown.
\textbf{Bottom row}: Learned vs true pendulum dynamics using matrix variate Gaussian Process regression
}
\label{fig:CPS}
    \label{fig:learning-accuracy}
\end{figure}

We evaluate the proposed approach on a pendulum with mass $m$ and length $l$ with state $\bfx = [\theta, \omega]$ and control-affine dynamics $f(\bfx) = [\omega, -\frac{g}{l}\sin(\theta)]$ and $g(\bfx) = [0, \frac{1}{ml}]$ as depicted in Fig~\ref{fig:CPS}. A safe set is chosen as the complement of a radial region $[\theta_c -\Delta_{col}, \theta_c + \Delta_{col}]$ that needs to be avoided. The controller knows a priori that the system is control-affine with relative degree two, but it is not aware of $f$ and $g$. The control barrier function is thus $h(\bfx) = \cos(\Delta_{col}) - \cos(\theta - \theta_c)$. We formulate a quadratically constrained quadratic program as in~\eqref{progg!!5544i} for $r=2$. We specify a task requiring the pendulum to track a reference control signal $\bfu_0$ and specify the optimization objective as $(\bfu_k-\textbf{u}_0)^\top Q(\bfu_k-\textbf{u}_0)$. We initialize the system with parameters $\theta_0=75^\circ$, $\omega_0=-0.01$, $\tau=0.01$, $m=1$, $g=10$, $l=1$, $\theta_c=45$, $\Delta_{col}=22.5$. The system dynamics are approximated accurately (see Fig.~\ref{fig:learning-accuracy}) while the system remains in the safe region (see Fig.~\ref{fig:CPS}). An $\epsilon$-greedy exploration strategy is used to sample $\bfu_0 \in [-20, 20]$. We use an exponentially decreasing $\epsilon$-greedy scheme going from $1$ to $0.01$ in $100$ steps. Negative control inputs get rejected by the CBF-based constraint, while positive inputs allow the pendulum to bounce back from the unsafe region.



%% file: tex/conculsion.tex
\section{Conclusion}\label{sec:conc}
Allowing artificial systems to safely adapt their own models during online operation will have significant implications for their successful use in unstructured, changing real-world environments. This paper developed a Bayesian inference approach to approximate system dynamics and their uncertainty from online observations. The posterior distribution over the dynamics may be used to enforce probabilistic constraints that guarantee safe online operation with high probability. Our results offer a promising approach for controlling complex systems in challenging environments. Future work will focus on extending the self-triggering time analysis to systems with higher relative degree and on applications of the proposed approach to real robot systems.


%% file: tex/note11.tex
We use $\otimes$ for the Kronecker product, and $\erf$ for the Gauss error
function. $\partial \mathcal{C}$ denotes the boundry of the set $\mathcal{C}$.
Let $\textit{vec}(\bfX) \in \mathbb{R}^{nm}$ be the vectorization of $\bfX \in
\R^{n \times m}$, obtained by stacking the columns of $\bfX$. For the functions $h : \real \rightarrow
\real^n$  and $f : \real^n \rightarrow
\real^n$ the Hessian and Jacobian are defined as follows
%
\newcommand{\p}{\partial}
\begin{align}
  H_{xx} h(\bfx) &:= \begin{bmatrix}
    \frac{\p h}{\p x_1 \p x_1} & \dots & \frac{\p h}{\p x_1 \p x_n}
    \\
    \vdots & & \vdots
    \\
    \frac{\p h}{\p x_n \p x_1} & \dots & \frac{\p h}{\p x_n \p x_n}
  \end{bmatrix},
                                       &
    J_\bfx f(\bfx) &:= \begin{bmatrix}
    \frac{\p f_1}{\p x_1} & \dots & \frac{\p f_1}{\p x_n}
    \\
    \vdots & & \vdots
    \\
    \frac{\p f_n}{\p x_1} & \dots & \frac{\p f_n}{\p x_n}
    \end{bmatrix}.
\end{align}%
The following parameters are useful for our analysis in Sec.~\ref{sec:exp-cbf-rel-deg-2} 
  \begin{equation}
    \eta(\bfx) := \begin{bmatrix} h(\bfx) \\ \Lie_fh(\bfx)
      \\ \vdots\\ \Lie_f^{(r-1)}h(\bfx) \end{bmatrix},
    \quad
    \mathcal{F} := \begin{bmatrix} 0  & 1 & \dots & 0 \\
      \vdots & \vdots &  & \vdots \\
      0 & 0 & \dots  & 1 \\
      0 & 0 & \dots & 0
    \end{bmatrix},
    \quad
    \mathcal{G} := \begin{bmatrix} 0 \\ \vdots \\ 1 \end{bmatrix}.
  \end{equation}

%% file: tex/MGVresults.tex
\begin{defn}
The Matrix Variate Gaussian (MVG) distribution is a three-parameter distribution $\calM\calN(\bfM,\bfA,\bfB)$ describing a random matrix $\bfX \in \mathbb{R}^{n \times m}$ with probability density function:
\begin{equation}
p(\bfX; \bfM, \bfA, \bfB) := \frac{\exp\prl{ -\frac{1}{2}\tr\brl{\bfB^{-1}(\bfX-\bfM)^\top \bfA^{-1}(\bfX-\bfM)}}}{(2\pi)^{nm/2}\det(\bfA)^{m/2}\det(\bfB)^{n/2}}
\end{equation}
where $\bfM \in \mathbb{R}^{n \times m}$ is the mean, and $\bfA \in \mathbb{R}^{n \times n}$, $\bfB \in \mathbb{R}^{m \times m}$ encode the covariance matrix of the rows and columns of $\bfX$, respectively.
\end{defn}

The two following lemmas are used to derive our results in Sec.~\ref{learningsec:13452}.

\begin{lemma}[{Vectorization of MVG~\cite{StructuredPBP}}]
  \label{lemma:mvg2gaussian}
Let $\bfX$ follow a MVG distribution $\calM\calN(\bfM,\bfA,\bfB)$. Then,
\begin{align}
\vect(\bfX) &\sim \cal{N}\prl{ \textit{vec}(\bfM), \bfB\otimes\bfA} 
\end{align}
\end{lemma}

\begin{lemma}[{Linear form on MVG}]
  \label{lemma:product-mvg}
Let $\bfX$ follow a MVG distribution $\calM\calN(\bfM,\bfA,\bfB)$ and let $\bfC \in \mathbb{R}^{d \times n}$ and $\bfD \in \mathbb{R}^{m \times d}$. Then,
\begin{equation}
\label{sunerrww233!}
\begin{aligned}
  \bfC \bfX &\sim \calM\calN(\bfM,\bfC\bfA\bfC^\top,\bfB),
  &
\bfX \bfD &\sim \calM\calN(\bfM,\bfA,\bfD^\top\bfB\bfD),
\end{aligned}
\end{equation}
and
\begin{align}
\label{rrr223!!!}
\cov(\vect(\bfC \bfX), \vect(\bfX)) &= \bfB \otimes \bfC \bfA
  &
\cov(\vect(\bfX \bfD), \vect(\bfX)) &= \bfD^\top \bfB \otimes \bfA
\end{align}
\end{lemma}
We refer the reader for the proof of~\eqref{sunerrww233!} to the paper~\cite{StructuredPBP}, and the
 the proof of~\eqref{rrr223!!!} is included in the appendix~\ref{proof:mvg-cov}.

\subsubsection{Comparison of Matrix Variate GP with alternatives}
Here we discuss two alternative approaches to infer a posterior over $F(\bfx)$, and mention the benifits of the proposed MVG framework with respect to them. The first alternative approach is to develop a decoupled GP regression per system dimension, which, unlike our MVG approach, does not model the dependencies between different components of $f(\bfx)$ and $g(\bfx)$. Moreover, the inference computational complexity of our MVG approach is $O((1+m)^3 k^2) + O(k^3)$  while the same for independent GP is $O((1+m)k^2) + O(k^3)$, where $k$ is the number of data points and $m$ is the control dimension. Thus for $k >> m$ which is common, the MVG has similar computational complexity as the independent GP approach but provides greater inference flexibility.
The second alternative approach is the Coregionalization models~\cite{alvarez2012kernels}, where we can simplify the covariance structure by assuming that $\bfK_0(\bfx,\bfx') = \bfSigma \knl_0(\bfx,\bfx')$ decomposes into a scalar state-dependent kernel $\knl_0(\state, \state')$ and an output-dimension-dependent covariance matrix $\bfSigma \in \R^{n(1+m) \times (1+m)n}$. Our MVG approach is more efficient as for systems with large state or control dimensions, learning the $(1+m)^2n^2$ parameters of $\bfSigma$ may require large amounts of training data. Moreover, the nice matrix-times-scalar-kernel structure is not preserved in the posterior using these Coregionalization models.

%% file: tex/moreonECBF.tex
If $K_\alpha$ satisfies the properties stated in in the next theorem, any control policy $\bfu = \pi(\bfx)$ that ensures $\mbox{CBC}^{(r)} \ge 0$, renders the dynamics~\eqref{eq:system_dyanmics} safe with respect to set $\mathcal{C}$.
\begin{theorem}[Designing $K_\alpha$~\cite{nguyen2016exponential,cbf}]
  Function $h(\bfx)$ is Exponential CBF if $K_\alpha$ is chosen with following
  conditions.
  \begin{enumerate}
    \item $\mathcal{F} - \mathcal{G}K_\alpha$ is Hurwitz and total negative
      (resulting in negative real poles).
    \item The eigenvalues $\lambda_i$ of the system \eqref{eq:traverse-system}
      satisfy
      $\lambda_i (\mathcal{F} - \mathcal{G}K_\alpha) \ge
      -\frac{\dot{\nu}_{i-1}(\bfx_0)}{\nu_{i-1}(\bfx_0)}$, where $\nu_i(\bfx)$
      is recursively defined as $\nu_0(\bfx) := h(\bfx)$, $\nu_i(\bfx) := \dot{\nu}_{i-1} +
      p_i\nu_{i-1}(\bfx)$, with $\{p_i\}_{i=1}^r$  as
    the $r$ roots of the characteristic polynomial of the matrix
    $\mathcal{F} - \mathcal{G}K_\alpha$.
  \end{enumerate}
\end{theorem}

For relative degree one systems $K_\alpha \eta(\bfx)$ reduces to $\alpha h(\bfx)$ with $\alpha>0$. Thus, $CBC^{(1)}$, the safety condition based on ECBF, is equivalent to CBC in Def.~\ref{def:cbf} when  $\alpha$, the extended class $K_\infty$ function, is linear.

%% file: tex/mean-and-variance-of-cbf-2/proof-mvg-cov.tex
\subsubsection{Matrix Variate Gaussian Covariance (Lemma~\ref{lemma:product-mvg})}
\label{proof:mvg-cov}

Let $\bfY := \bfX - \bfM$, such that $\bfY \sim \mathcal{MVG}(0, \bfA, \bfB)$.
Let $\bfY = [ \bfy_1, \dots, \bfy_m ]$ where $\bfy_i \in \R^n$ are column
vectors of $\bfY$.
\begin{align}
  \E[\vect(\bfY)\vect(\bfY)^\top]
  &=
  \E\left[\begin{bmatrix}
    \bfy_1\bfy_1^\top & \dots  & \bfy_1\bfy_m^\top
    \\ \vdots & & \vdots \\
    \bfy_m\bfy_1^\top & \dots  & \bfy_m\bfy_m^\top
    \end{bmatrix}\right]
                                 =  \bfB \otimes \bfA =
  \begin{bmatrix}
    b_{11}\bfA & \dots  & b_{1m}\bfA
    \\ \vdots & & \vdots \\
    b_{m1}\bfA & \dots  & b_{mm}\bfA
    \end{bmatrix}
\end{align}%

Note that $\bfC\bfY = [\bfC\bfy_1, \dots, \bfC\bfy_m]$.
\begin{align}
  \cov(\vect(\bfC\bfY), \vect(\bfY))
  &= \E[\vect(\bfC\bfY)\vect(\bfY)^\top]
  =
  \E\left[\begin{bmatrix}
    \bfC\bfy_1\bfy_1^\top & \dots  & \bfC\bfy_1\bfy_m^\top
    \\ \vdots & & \vdots \\
    \bfC\bfy_m\bfy_1^\top & \dots  & \bfC\bfy_m\bfy_m^\top
    \end{bmatrix}\right]
  \\
  &=
  \begin{bmatrix}
    b_{11}\bfC \bfA & \dots  & b_{1m}\bfC \bfA
    \\ \vdots & & \vdots \\
    b_{m1}\bfC \bfA & \dots  & b_{mm}\bfC \bfA
    \end{bmatrix}
                               =  \bfB \otimes \bfC \bfA 
\end{align}%

Second part can be proved by noticing that,
\begin{align}
\bfX &\sim \mathcal{MVG}(\bfM, \bfA, \bfB)
  \implies \bfX^\top \sim \mathcal{MVG}(\bfM, \bfB, \bfA)
  \\
  &\implies \cov(\vect(\bfD^\top \bfX^\top),\vect(\bfX^\top)) = \bfA \otimes \bfD^\top \bfB
  \\
     &\implies \cov(\vect(\bfX \bfD),\vect(\bfX)) = \bfD^\top \bfB \otimes \bfA
\end{align}%

%% file: tex/mean-and-variance-of-cbf-2/matrix-variate-gp.tex
\subsubsection{Matrix-variate Gaussian Process (Eq~\eqref{eq:mvg-posterior})}
\label{proof:matrix-variate-gp}
We start by noticing that the Kronecker product satisfies the  following properties.  
\begin{align}
  (P \otimes Q)(R \otimes S) &= (PR) \otimes (QS) \label{eq:kron-mixed-prod}
  \\
(P \otimes Q)^{-1} &=  P^{-1} \otimes Q^{-1} \label{eq:kron-inv}
  \\
  ( Q + R )\otimes P &= Q \otimes  P + R \otimes P. \label{eq:kron-bilinear}
\end{align}%

We use the above properties of kronecker product to show how $\bfA$ can be
factorized out of variance and removed from mean of GP posterior. First we
consider the term that appears in both mean and variance,
\begin{align}
  \Bigl( [& \CtDat^\top \bfB_0(\bfX_{1:k}, \bfx_*) \otimes \bfA \Bigr)^\top
\Bigl([ \CtDat^\top \bfB_0(\bfX_{1:k}, \bfX_{1:k}) \CtDat ] \otimes \bfA \Bigr)^{-1}
  \\
  &= \Bigl( [ \CtDat^\top \bfB_0(\bfX_{1:k}, \bfx_*) \otimes \bfA \Bigr)^\top
\Bigl([ \CtDat^\top \bfB_0(\bfX_{1:k}, \bfX_{1:k}) \CtDat ]^{-1} \otimes \bfA^{-1} \Bigr)
    & \text{ using \eqref{eq:kron-inv} }
  \\
  &= \Bigl( [ \CtDat^\top \bfB_0(\bfX_{1:k}, \bfx_*)^\top
[ \CtDat^\top \bfB_0(\bfX_{1:k}, \bfX_{1:k}) \CtDat ]^{-1}\Bigr) \otimes (\bfA \bfA^{-1})
    & \text{ using \eqref{eq:kron-mixed-prod} }
  \\
  &= \Bigl(  \bfB_0(\bfX_{1:k}, \bfx_*)^\top \CtDat
    [ \CtDat^\top \bfB_0(\bfX_{1:k}, \bfX_{1:k}) \CtDat ]^{-1}
    \Bigr) \otimes I_n
\end{align}%

\newcommand{\tdbfX}{\tilde{\bfX}}
\newcommand{\tdbfx}{\tilde{\bfx}}
Now we consider the mean,
\begin{align}
  &\vect(\bfM_k(\bfx_*))
   - \vect(\bfM_0(\bfx_*))
  \\
  &=
  \Bigl( [ \CtDat^\top \bfB_0(\bfX_{1:k}, \bfx_*) \otimes \bfA \Bigr)^\top
\Bigl([ \CtDat^\top \bfB_0(\bfX_{1:k}, \bfX_{1:k}) \CtDat ] \otimes \bfA \Bigr)^{-1}
\vect(\StDtDat - \bfM_0(\bfx_*))
  \\
  &=
\Biggl(\underbrace{\Bigl[  \bfB_0(\bfX_{1:k}, \bfx_*)^\top \CtDat
    [ \CtDat^\top \bfB_0(\bfX_{1:k}, \bfX_{1:k}) \CtDat ]^{-1}
    \Bigr]}_{P} \otimes I_n\Biggr)\vect(\tdbfX)
  \\
  &=
  \begin{bmatrix}
P_{1,1} I_n & \dots & P_{1,d} I_n \\ 
\vdots &  & \vdots \\
P_{1+m,1} I_n & \dots & P_{1+m,d} I_n
\end{bmatrix}
  \begin{bmatrix}
    \tdbfx_1 \\ \vdots \\ \tdbfx_d
    \end{bmatrix}
  \\
  &=
  \begin{bmatrix}
P_{1,1} \tdbfx_1 & \dots & P_{1,d} \tdbfx_d \\ 
\vdots &  & \vdots \\
P_{1+m,1} \tdbfx_1 & \dots & P_{1+m,d} \tdbfx_d
\end{bmatrix}
  \\
  &=
  \begin{bmatrix}
\tdbfX P^\top_{1,1:d} \\
\vdots \\
\tdbfX P^\top_{1+m,1:d}
\end{bmatrix}
  \\
  \bfM_k(\bfx_*)& - \bfM_0(\bfx_*)
  = \tdbfX P^\top
  \\
&= (\StDtDat - \bfM_0(\bfx_*))
    [ \CtDat^\top \bfB_0(\bfX_{1:k}, \bfX_{1:k}) \CtDat ]^{-\top}
        \CtDat^\top\bfB_0(\bfX_{1:k}, \bfx_*)
\end{align}%
Note that $\bfA$ does not affect the mean.

Now we consider the second term in the variance,
\begin{align}
  \Bigl( [& \CtDat^\top \bfB_0(\bfX_{1:k}, \bfx_*) \otimes \bfA \Bigr)^\top
\Bigl([ \CtDat^\top \bfB_0(\bfX_{1:k}, \bfX_{1:k}) \CtDat ] \otimes \bfA \Bigr)^{-1}
            \Bigl( [ \CtDat^\top \bfB_0(\bfx_*, \bfX_{1:k})] \otimes \bfA \Bigr)
  \\
  &= \Biggl[
  \Bigl(  \bfB_0(\bfX_{1:k}, \bfx_*)^\top \CtDat
    [ \CtDat^\top \bfB_0(\bfX_{1:k}, \bfX_{1:k}) \CtDat ]^{-1}
    \Bigr) \otimes I_n \Biggr] 
  \Bigl( [ \CtDat^\top \bfB_0(\bfx_*, \bfX_{1:k})] \otimes \bfA \Bigr)
  \\
  &=
  \Biggl[
   \bfB_0(\bfx_*, \bfX_{1:k})^\top \CtDat
    \Bigl( \CtDat^\top \bfB_0(\bfX_{1:k}, \bfX_{1:k}) \CtDat \Bigr)^{-1}
   \CtDat^\top \bfB_0(\bfx_*, \bfX_{1:k})
  \Biggr] \otimes \bfA
    ~~~~\text{ using \eqref{eq:kron-mixed-prod}}
\end{align}

Now, we can compute the variance of the posterior,
\begin{align}
  &\bfB_k(\bfx_*, \bfx'_*) \otimes \bfA=
  \\
  &\bfB_0(\state_*, \state'_*) \otimes \bfA 
  - \Biggl[
   \bfB_0(\bfx_*, \bfX_{1:k})^\top \CtDat
    \Bigl( \CtDat^\top \bfB_0(\bfX_{1:k}, \bfX_{1:k}) \CtDat \Bigr)^{-1}
   \CtDat^\top \bfB_0(\bfx'_*, \bfX_{1:k})
  \Biggr] \otimes \bfA
  \\
  &= \Biggl[\bfB_0(\state_*, \state'_*)
  - 
   \bfB_0(\bfx_*, \bfX_{1:k})^\top \CtDat
    \Bigl( \CtDat^\top \bfB_0(\bfX_{1:k}, \bfX_{1:k}) \CtDat \Bigr)^{-1}
   \CtDat^\top \bfB_0(\bfx'_*, \bfX_{1:k})
  \Biggr] \otimes \bfA
\end{align}%

%% file: tex/CBCerror.tex
\subsubsection{Proof of Lemma~\ref{lem:NRcalculation} (The posterior of CBC)}
We start by re-writing the definition of CBC as follows. 
\begin{align}
  \mbox{CBC}_k&=
                                  \nabla_\bfx h(\bfx_k)^\top\begin{bmatrix}f(\bfx_k) &g(\bfx_k)\end{bmatrix}\begin{bmatrix}
1
\\
\bfu_k  
\end{bmatrix}+(\alpha \circ h)(\bfx_k).
\end{align}
Since $\begin{bmatrix}
f(\bfx) & g(\bfx)
\end{bmatrix}$ are jointly Gaussian,
their linear combination is also Gaussian. We further notice conditioned on $\bfx_k$ and $\bfu_k$ we have
\begin{align}
   \E[f(\bfx_k)+g(\bfx_k)\bfu_k ]
&= \barf^*_k(\bfx_k)+\barg^*_k(\bfx_k)\bfu_k
  \\
 \Var{\left(\begin{bmatrix}f(\bfx) &g(\bfx)\end{bmatrix}\begin{bmatrix}
1
\\
\bfu_k  
\end{bmatrix}\right)} &= 
 \begin{bmatrix}
1
\\
\bfu_k  
\end{bmatrix}^\top\Var{\left(\begin{bmatrix}f(\bfx) &g(\bfx)\end{bmatrix}\right)}\begin{bmatrix}
1
\\
\bfu_k  
\end{bmatrix}
\end{align}
and the result follows.

%% file: tex/KnonwLipshitzsefltriggering.tex
\subsubsection{Proof of Theorem~\ref{ir5o45t!!!} (self-triggered control)}
Since the program~\eqref{prog:CBF-CLF:unknown} have  solutions at the triggering times we know $\mathbb{P}(\mbox{CBC}_k \ge \zeta | \bfx_k,\bfu_k) \ge \tilde{p}_k$ and $\mathbb{P}(\mbox{CBC}_{k+1} \ge \zeta| \bfx_k,\bfu_k) \ge \tilde{p}_{k+1}$. Thus, if we prove condition on $\mbox{CBC}_k \ge \zeta$, $\mbox{CBC}_{k+1} \ge \zeta$, and the event~\eqref{eq:smoth23}, we have $\mbox{CBC}(s+t_k) \ge 0$, for all $s \in [0, \tau_k)$, the result follows.

Using Cauchy-Schwarz inequality, and the Lipschitz assumptions~\eqref{eq:smoth23} and~\eqref{htym6!7uytf}, and Proposition~\ref{self-trige3Lip}, for all $s \in [0, \Delta_k)$, we have
\begin{align}\label{nigrer45654e!}
    |\mbox{CBC}(x(s+t_k))-\mbox{CBC}_k| \le 
   \left(\sup_{s \in [0, \Delta_k)}\|\nabla h(x(s+t_k))\|L_k+L_{\alpha \circ h}\right)\overline{r}_k(s).
\end{align}
Thus, using~\eqref{nirn3453456543!} and~\eqref{defnirrrt5654}, we notice the right-hand side of~\eqref{nigrer45654e!} is upper-bounded by
\begin{align}\label{nigrer45654e!345qw3}
    \frac{\chi_kL_k+L_{\alpha \circ h}}{L_k}\|\dot{\bfx}_k\|\left(e^{L_ks}-1\right).
\end{align}
The result follows by noticing for $s=\tau_k$~\eqref{nigrer45654e!345qw3} is less than or equal to $\zeta$.

%% file: tex/cbc-r-mean-and-variance-affine-and-quadratic.tex
\subsubsection{Proof of Proposition~\ref{prob:VarUquadrac24562} ($\mbox{CBC}^{(r)}$ mean is affine and its variance is quadratic)
}
\label{proof:cbc-r-mean-affine-var-quadratic}


This proof uses the linearity of expectation, 
\newcommand{\CBCr}{\mbox{CBC}_k^{(r)}}
\begin{align}
    \E[\mbox{CBC}^{(r)}(\bfx, \bfu)] 
    &=
    \E[\Lie_f^{(r)}h(\bfx)] + \E[\Lie_g\Lie_f^{(r-1)}h(\bfx)]\bfu+K_\alpha \E[\eta(\bfx)]
    \\
    &=
    \E[\Lie_g\Lie_f^{(r-1)}h(\bfx)]\bfu
    +\E[\Lie_f^{(r)}h(\bfx)] + K_\alpha \E[\eta(\bfx)]
    \\
    \Var[\mbox{CBC}^{(r)}(\bfx, \bfu)] &= 
    \Var\big[(\Lie_f^{(r)}h(\bfx) + K_\alpha \eta(\bfx))\big]
    \nonumber\\
    &\qquad+ \Var\big[\Lie_g\Lie_f^{(r-1)}h(\bfx)\bfu\big]
    \nonumber\\
    &\qquad+ \cov[( \Lie_f^{(r)}h(\bfx) 
    + K_\alpha \eta(\bfx)), (\Lie_g\Lie_f^{(r-1)}h(\bfx)  \bfu) ] 
    \nonumber\\
    &\qquad+ \cov[ (\Lie_g\Lie_f^{(r-1)}h(\bfx)  \bfu),
    ( \Lie_f^{(r)}h(\bfx)  + K_\alpha \eta(\bfx)) ] .
    \\
    &=
    \bfu^\top \Var\big[\Lie_g\Lie_f^{(r-1)}h(\bfx)\big]\bfu
    \nonumber\\
    &\qquad+ 
    2\cov[ (\Lie_g\Lie_f^{(r-1)}h(\bfx) ),
    ( \Lie_f^{(r)}h(\bfx)  + K_\alpha \eta(\bfx)) ]^\top  \bfu .
    \nonumber\\
    &\qquad+ 
    \Var\big[(\Lie_f^{(r)}h(\bfx) + K_\alpha \eta(\bfx))\big]
\end{align}
Each of the term in $\Var(\CBCr)$ is at most quadratic in $\bfu$, hence $\Var(\CBCr)$ is quadratic in $\bfu$.

\subsubsection{Proof of Proposition~\ref{prop:456723o433302}}
\label{proof:cbc-r-quad-prog}
Unlike $\mbox{CBC}_k$ for relative degree one, the distribution of $\mbox{CBC}_k^{(r)}$ is not a Gaussian Process for $r \ge 2$. 
Hence instead of analytically computing the probability distribution, we use Cantelli's inequality to bound the mean and variance of $\CBCr$ for any scalar $\lambda <0 $, we have
\begin{align}
\label{Caneeekee44!}
    \mathbb{P}\left(\mbox{CBC}_k^{(r)} \ge \E[\mbox{CBC}_k^{(r)}]+\lambda | \bfx_k,\bfu_k\right) \ge
    1 - \frac{\Var[\mbox{CBC}_k^{(r)}]}{\Var[\mbox{CBC}_k^{(r)}]+\lambda^2}.
\end{align}%
Since we want the probability to be greater than $\tilde{p}_k$, we ensure its lower bound is greater than $\tilde{p}_k$, i.e.
\begin{align}
    1 - \frac{\Var[\mbox{CBC}_k^{(r)}]}{\Var[\mbox{CBC}_k^{(r)}]+\lambda^2} \ge \tilde{p}_k
\end{align}
The terms can be rearranged into,
\begin{align}
(1-\tilde{p}_k)(\Var[\mbox{CBC}_k^{(r)}]+\lambda^2) &\ge \Var[\mbox{CBC}_k^{(r)}]
\\
\lambda^2 &\ge \frac{\tilde{p}_k}{1-\tilde{p}_k}\Var[\mbox{CBC}_k^{(r)}]
\end{align}
If $\E[\mbox{CBC}_k^{(r)}] \ge \zeta$ we can substitute  $\lambda= -(\E[\mbox{CBC}_k^{(r)}]-\zeta) < 0$, thus
\begin{align}
(\E[\mbox{CBC}_k^{(r)}]-\zeta)^2 &\ge \frac{\tilde{p}_k}{1-\tilde{p}_k}\Var[\mbox{CBC}_k^{(r)}]
\quad \text{ and } \quad \E[\mbox{CBC}_k^{(r)}]-\zeta \ge 0
\nonumber\\
&\implies  \mathbb{P}\left(\mbox{CBC}_k^{(r)} \ge \zeta | \bfx_k,\bfu_k\right) \ge \tilde{p}_k
\label{eq:cbr-r-ineq}
\end{align}
Substituting the constraint in \eqref{Caneeekee44!}, we get \eqref{progg3456755!!}.
\begin{align}
\min_{\bfu_k
} \;\;& \bfu_k^\top Q\bfu_k
&
\text{s.t.}~~~~ &\E[\mbox{CBC}_k^{(r)}] - \zeta \ge 0~~~~
\text{and}~~
(\E[\mbox{CBC}_k^{(r)}] -\zeta)^2 \ge \frac{\tilde{p}_k}{1-\tilde{p}_k} \Var[\mbox{CBC}_k^{(r)}]
\end{align}

%% file: tex/mean-and-variance-of-cbf-2.tex
\subsubsection{Mean and Variance of $\CBCtwo$~\ref{sec:exp-cbf-rel-deg-2}}
\label{proof:mean-var-of-cbc}
\newcommand{\mCBCtwo}{\E[\CBCtwo(\bfx; \bfu)]}
\newcommand{\VCBCtwo}{\Var(\CBCtwo(\bfx; \bfu))} 
\newcommand{\postf}{\bff_k}
\newcommand{\LieOne}{\Lie_{f_k}}
\newcommand{\LieTwo}{\Lie_{\postf}^{(2)}}
\newcommand{\LieTwoF}{\Lie_{f_k}^{(2)}}
\newcommand{\LieTwoG}{\Lie_{g_k} \LieOne}
\newcommand{\bfzero}{\mathbf{0}}
\newcommand{\mutdf}{{\mu}_{f_k}}
\newcommand{\ktdf}{{\knl}_{f_k}}
\newcommand{\mutdgu}{{\mu}_{g_k}}
\newcommand{\ktdgu}{{\knl}_{g_k }}
\newcommand{\mutdbfu}{{\mu}_{\bff_k}}
\newcommand{\ktdbfu}{{\knl}_{\bff_k }}
\newcommand{\ktdfgu}{{\knl}_{f_k,g_k}}
\newcommand{\ktdfbfu}{\bfb_k^\top}

The mean, variances and covariances
of different components are given by (using~Lemma~\ref{lemma:product-mvg}),
\begin{align}
  \mutdf(\bfx) &:=  \mDynAffs(\bfx) [1, \bfzero_m]^\top,
  & \Var(f_k(\bfx), f_k(\bfx')) &= \underbrace{[1, \bfzero_m] \bfBs(\bfx, \bfx') [1, \bfzero_m]^\top}_{ =: \ktdf(\bfx, \bfx') \in \R} \bfA
                                        \label{eq:mvg-posterior-f}
  \\
  \mutdgu(\bfx)\ctrlaff &:= \mDynAffs(\bfx) [0, \bfu^\top]^\top,
  & \Var(g_k(\bfx)\bfu, g_k(\bfx')\bfu') &= \underbrace{[0, \bfu^\top] \bfBs(\bfx, \bfx') [0, \bfu'^\top]^\top}_{=: \ctrlaff'^\top\ktdgu(\bfx, \bfx')\ctrlaff' \in \R} \bfA
  \\
  &
  & \Var(\dynAff(\bfx)\ctrlaff, \dynAff_k(\bfx') \ctrlaff) &= \ctrlaff^\top \bfBs(\bfx, \bfx') \ctrlaff \bfA
                                                    \label{eq:mvg-posterior-bfu}
  \\
               && \cov(f_k(\bfx), \dynAff_k(\bfx')\ctrlaff')
  &= \underbrace{[1, \bfzero_m] \bfBs(\bfx, \bfx')}_{\bfb_k(\bfx, \bfx')^\top}\ctrlaff \bfA.
\end{align}%
Note that all the variances are expressed as a scalar kernel function times $\bfA$.

\newcommand{\tdmu}{\hat{\mu}}
\newcommand{\tdbfk}{\hat{\bfk}}
\newcommand{\hatf}{f_k}
\newcommand{\hDynAff}{\dynAff_k}
 With this new notation $\CBCtwo$ can be written as,
\begin{align}
\CBCtwo(\bfx; \bfu) &= \grad \LieOne h(\bfx) \dynAff_k \ctrlaff + K_\alpha [h(\bfx), \Lie_{f_k}h(\bfx)]^\top.
\end{align}%

\begin{theorem}
\label{thm:CBC2-quanti34}
The mean and variance of $\mbox{CBC}_k^{(2)}$ for relative-degree 2 system ($\Lie_g
h(x) = 0$) are,
%
\begin{equation}\begin{split}%
\mCBCtwo
&= 
\E[ \grad \LieOne h(\bfx) \dynAff_k ] \ctrlaff
  + K_{\alpha}\begin{bmatrix} h(\bfx)
   & \E[\LieOne h(\bfx)]
   \end{bmatrix}^\top
\\
\VCBCtwo
&=
\Var[ \grad \LieOne h(\bfx) \dynAff_k \ctrlaff ]
+ \Var[ K_{\alpha,2} \LieOne h(\bfx) ]
\\&\quad 
+ K_{\alpha,2} \cov(\LieOne h(\bfx), \grad \LieOne h(\bfx) \dynAff_k \ctrlaff )
\\&\quad
+ K_{\alpha,2} \cov(\grad \LieOne h(\bfx) \dynAff_k \ctrlaff, \LieOne h(\bfx))
\label{eq:mean-var-cbc}
\end{split}\end{equation}%
where each term in the above equations can be computed step by step by Algorithm~\ref{alg:Algo}.
%
\end{theorem}

We will compute the mean and variance of each of the term one by one.

We derive the mean and variance and bounds over $\CBCtwo$ for a Gaussian
Process $\dynAff(\bfx)\ctrlaff \sim \GP(\bfM_0(\bfx)\ctrlaff, \ctrlaff^\top \bfB_k(\bfx, \bfx')\ctrlaff)$.
Then condition
$\LieTwo h(\bfx)$ can be written as a quadratic form in the random
vectors $\bfz(\bfx; \bfu) := [\grad \LieOne h(\bfx)^\top, \dynAff(\bfx)\ctrlaff,
\LieOne  h(\bfx)]^\top$ ,
\begin{align}
  \LieTwoF(\bfx) + \LieTwoG(\bfx)\bfu
  := \grad \LieOne h(\bfx)^\top \dynAff(\bfx)\ctrlaff
   = \bfz(\bfx; \bfu)^\top
  \underbrace{\begin{bmatrix}
    0 & \frac{I_n}{2} &  0 \\
    \frac{I_n}{2} & 0 &  0 \\
    0 & 0 & 0
  \end{bmatrix}}_{=: \Lambda }
            \bfz(\bfx; \bfu)
  \label{eq:cbc-quadratic-form}
\end{align}%

The challenge is to compute the mean and variance of vector-variate Gaussian
process $\bfz(\bfx; \ctrlaff)$.
We compute it in three main steps, computation of mean and variance of (1) $\LieOne h(\bfx)$, (2) $\grad \LieOne h(\bfx)$ and (3) $\grad \LieOne h(\bfx) \dynAff(\bfx) \ctrlaff$.

\paragraph{Step 1: Mean and variance of $\LieOne h(\bfx)$}
We start by observing that $\LieOne h(\bfx) = \grad h(\bfx)^\top f_k(\bfx) $ is a
scalar value Gaussian process because $\grad h(\bfx)$ is deterministic,
\begin{equation}\begin{split}%
  \LieOne h &\sim \GP\left(\overline{\LieOne h} , \knl_{\LieOne h} \right)
  \\
  \text{where } \overline{\LieOne h} (\bfx) &= \grad h(\bfx)^\top \mutdf(\bfx)
  \\
  \knl_{\LieOne h}(\bfx, \bfx') &= \ktdf(\bfx, \bfx') \grad h(\bfx)^\top \bfA \grad h(\bfx').
  \\
  \cov(\dynAff(\bfx')\ctrlaff, \LieOne h(\bfx)) &= \bfb_k(\bfx, \bfx')^\top\ctrlaff\bfA\grad h(\bfx) 
  \label{eq:mean-var-lie-grad-h-f}
\end{split}\end{equation}%

\paragraph{Step 2: Mean and variance of $\grad_\bfx \LieOne h(\bfx)$}
\begin{lemma}[Differentiating a Gaussian Process]
  \label{lemma:differentiating-gp}
  Let $q$ be a scalar valued Gaussian Process with mean function
$\mu : \R^n \to \R$ and kernel function $\knl : \R^n \to \R$, then $\grad q$ is a vector-valued Gaussian
Process,
\begin{align}
  \grad q &\sim \GP(\grad \mu, K_{\grad q}) \\
  \text{where }&  K_{\grad q}(\bfx, \bfx') = H_{\bfx, \bfx'} \knl(\bfx, \bfx')
                                                                                     \\
&\cov(\grad q(\bfx), q(\bfx')) = \grad_{\bfx} \knl(\bfx, \bfx').
\end{align}

If $s$ is another random process whose finite covariance with $q$ is known to be
$\cov_{q,s}(\bfx, \bfx')$, then
\begin{align}
\cov(\grad q(\bfx), s(\bfx')) = \grad_\bfx \cov_{q,s}(\bfx, \bfx').
\end{align}%
For proof see Section~\ref{proof:differentiating-gp}. Note that the expressions for mean, variance and covariance are valid for any random process, even if not Gaussian.
\end{lemma}%

Using Lemma~\ref{lemma:differentiating-gp} $\grad \LieOne h(\bfx)$ is a
vector-variate Gaussian Process,
\begin{equation}\begin{split}%
  \grad \LieOne h &\sim \GP(\overline{\grad \LieOne h}, K_{\grad \LieOne h})
  \\
  \text{ where }&
  \overline{\grad \LieOne h}(\bfx) = \grad_\bfx \overline{\LieOne h} (\bfx)
  \\
  &K_{\grad \LieOne h}(\bfx, \bfx') = H_{\bfx,\bfx'}  \knl_{\LieOne h}(\bfx, \bfx')
    \\
    & \cov(\grad \LieOne h(\bfx), \LieOne h(\bfx)) = \grad_\bfx  \knl_{\LieOne h}(\bfx, \bfx')
    \\
    & \cov(\grad \LieOne h(\bfx), \dynAff(\bfx')\ctrlaff) = \grad_\bfx  \cov(\LieOne h(\bfx), \dynAff(\bfx')\ctrlaff))
\label{eq:mean-var-grad-lie-grad-h-f}
\end{split}\end{equation}%

The above derivatives can be expanded as,
\begin{equation}\begin{split}%
  \overline{\grad \LieOne h}(\bfx) &= H_{\bfx,\bfx}h(\bfx) \mutdf(\bfx) + J^\top_\bfx \mutdf(\bfx) \grad h(\bfx)
  \\
  K_{\grad \LieOne h}(\bfx, \bfx')
&= H_{\bfx, \bfx'}\ktdf(\bfx, \bfx') \grad h(\bfx)^\top \bfA \grad h(\bfx')
+ \grad h(\bfx')\bfA^{\top}  H_{xx} h(\bfx) \grad_\bfx' \ktdf(\bfx, \bfx')^\top
\\
&\quad
+ \grad_\bfx \ktdf(\bfx, \bfx') \grad h(\bfx)^\top \bfA H_{xx} h(\bfx')
\\
&\quad
+ \ktdf(\bfx, \bfx') H_{xx} h(\bfx) \bfA H_{xx} h(\bfx')
    \; \text{(See ~\ref{proof:differentiating-lie})}
    \\
  \cov(\grad \LieOne h(\bfx), \LieOne h(\bfx')) &= \grad_\bfx \ktdf(\bfx, \bfx') \grad h(\bfx)^\top \bfA \grad h(\bfx') + \ktdf(\bfx, \bfx') \grad h(\bfx') \bfA^\top H_{xx} h(\bfx)
  \\
\cov(\grad \LieOne h(\bfx)), \dynAff(\bfx')\ctrlaff) &= 
     J_\bfx^\top \bfb_k(\bfx,\bfx') \ctrlaff \bfA \grad h(\bfx)^\top
    +  H^\top_{xx} h(\bfx)\bfA \bfb_k(\bfx,\bfx')^\top\ctrlaff
\label{eq:mean-var-grad-lie-grad-h-f-expanded}
\end{split}\end{equation}%

\paragraph{Step 3: Mean and variance of $\grad_\bfx \LieOne h(\bfx) \dynAff(\bfx) \ctrlaff$}

Now we have all terms to write the mean and covariance of the Gaussian Process
$\bfz(\bfx;\bfu) \sim \GP(\bar{\bfz}, \Sigma_{\bfz})$, where
\begin{align}
  &\bar{\bfz}(\bfx; \bfu) = 
  \begin{bmatrix}
  \overline{\grad \LieOne h}(\bfx)^\top & 
    (\bfM_k(\bfx)\ctrlaff)^\top &
    \overline{\LieOne h}(\bfx)\end{bmatrix}^\top
    \\
    &\scaleMathLine{
\Sigma_\bfz(\bfx, \bfx'; \bfu) =
    \begin{bmatrix}
    K_{\grad \LieOne h}(\bfx, \bfx') &
    \cov(\grad \LieOne h(\bfx), \dynAff(\bfx') \ctrlaff) &
    \cov(\grad \LieOne h(\bfx), \LieOne h(\bfx))
    \\
    \cov(\grad \LieOne h(\bfx), \dynAff(\bfx') \ctrlaff)^\top
    & \ctrlaff^\top \bfB_k(\bfx, \bfx')\ctrlaff \bfA
    & \cov(\LieOne h(\bfx), \dynAff(\bfx') \ctrlaff)^\top
    \\
    \cov(\grad \LieOne h(\bfx), \LieOne h(\bfx))^\top
    & \cov(\LieOne h(\bfx), \dynAff(\bfx') \ctrlaff)
    & \knl_{\LieOne h}(\bfx, \bfx')
    \end{bmatrix}
      }.
      \label{eq:mean-var-z}
\end{align}
Note that $\Sigma_\bfz$ is also quadratic in control with the term $\cov(\grad
\LieOne h(\bfx), \tdbff(\bfx'; \bfu))$ being affine in $\bfu$ and the quadratic
term $\ctrlaff^\top \ktdbfu(\bfx, \bfx')\ctrlaff$.

To compute the mean and variance of $\CBCtwo$ from equation
\eqref{eq:cbc-quadratic-form}, we need the following lemma.

\begin{lemma}[Mean and cumulants of Quadratic form]
  \label{lemma:quadratic-form}
  Let $\bfx$ a Gaussian random variable with 
mean $\bar{\bfx}$, variance $\Sigma$. Let $\Lambda$ be symmetric. 
The mean of the Quadratic form $\bfx^\top \Lambda \bfx$ can be computed by
  \begin{align}
    \E[\bfx^\top \Lambda \bfx] 
      = \bar{\bfx}^\top \Lambda \bar{\bfx} + \tr(\Lambda \Sigma ).
  \end{align}%
  The above mean is valid for any random variable with finite mean and variance.
  
 The $r$th cumulant for Gaussian Random Variables is given by~\citep[p55]{searle1971linear}
  \begin{align}
  \calK_r(\bfx^\top \Lambda \bfx) = 2^{r-1}(r-1)![\tr(\Lambda \Sigma)^r + r\bar{\bfx}^\top \Lambda (\Sigma \Lambda)^{r-1} \bar{\bfx}].
  \end{align}%
  And the covariance of the quadratic form with the original variable is given by,
  \begin{align}
      \cov(\bfx, \bfx^\top \Lambda \bfx) = 2\Sigma \Lambda \bar{\bfx}
  \end{align}
\end{lemma}%

We begin by noting that the variance of quadratic form is the second cumulant $r=2$,
\begin{align}
  \Var(\bfx^\top \Lambda \bfx) = 2\tr(\Lambda \Sigma)^2 + 4\bar{\bfx}^\top \Lambda \Sigma \Lambda \bar{\bfx}
\end{align}

Consider two Gaussian random vectors $\bfx$ and $\bfy$ with mean $\bar{\bfx}$ and $\bar{\bfy}$ respectively and variances $\Var(\bfx)$ and $\Var(\bfy)$ respectively.
Let the covariance between $\bfx$ and $\bfy$ be given by $\cov(\bfx, \bfy)$. 
We want to find out the mean and variance of $\bfx^\top \bfy$. First note that it can be written in quadratic form,
\begin{align}
\bfx^\top \bfy = \begin{bmatrix}\bfx^\top & \bfy^\top\end{bmatrix}
\begin{bmatrix} 0 & 0.5 I \\ 0.5 I & 0 \end{bmatrix}
\begin{bmatrix} \bfx \\ \bfy \end{bmatrix}
\end{align}
Hence the mean and variance of $\bfx^\top \bfy$ is given by,
\begin{align}
\E[\bfx^\top \bfy]
&= \bar{\bfx}^\top \bar{\bfy} + \tr(\cov(\bfx, \bfy))
\\
\Var(\bfx^\top \bfy)
&= 2\tr(\cov(\bfx, \bfy))^2 + \bfy^\top \Var(\bfx) \bfy + \bfx^\top \Var(\bfy) \bfx + 2\bfy^\top \cov(\bfx, \bfy) \bfx
  \\
\cov(\bfx, \bfx^\top \bfy)
  &= \cov(\bfx, \bfy)\bar{\bfx} + \Var(\bfx) \bar{\bfy}
\end{align}

Now we are equipped to compute mean and variance of $\grad \LieOne  h(\bfx)^\top \dynAff(\bfx)\ctrlaff$:
\begin{align}
  \E[\grad \LieOne  h(\bfx)^\top \dynAff(\bfx)\ctrlaff ]
  &= 
  \E[\grad \LieOne h(\bfx)]\bfM_k(\bfx)\ctrlaff
  + \tr(\grad[\cov( \LieOne h(\bfx), \dynAff_k(\bfx)\ctrlaff))])
  \label{eq:lie-2-mean-app}\\
  &= \bigl(
      H_{\bfx,\bfx}h(\bfx) \mutdf(\bfx) + J^\top_\bfx \mutdf(\bfx) \grad h(\bfx)
    \bigr)
    \bfM_k(\bfx)\ctrlaff
    \nonumber\\
     &\quad + \tr(J_\bfx^\top \bfb_k(\bfx,\bfx') \ctrlaff \bfA \grad h(\bfx)^\top )
    +  \tr( H^\top_{xx} h(\bfx)\bfA \bfb_k(\bfx,\bfx')^\top\ctrlaff )
\end{align}
\begin{align}
  \Var[\grad \LieOne  h(\bfx)^\top \dynAff(\bfx)\ctrlaff ]
  &= 
  2\tr(\grad \cov(\LieOne h(\bfx), \dynAff_k(\bfx) \ctrlaff)))^2
    \nonumber\\
  &+ 2 \grad \E[\LieOne h(\bfx)] \grad \cov(\LieOne h(\bfx), \dynAff(\bfx) \ctrlaff) \E[\dynAff(\bfx)\ctrlaff]
    \nonumber\\
  &+ \grad \E[\LieOne h(\bfx)]^\top \ctrlaff^\top\bfB(\bfx, \bfx)\ctrlaff\bfA \grad \E[\LieOne h(\bfx)]
    \nonumber\\
  &+  [\bfM_k(\bfx)\ctrlaff]^\top H_{xx} \Var[\LieOne h(\bfx)] \bfM_k(\bfx)\ctrlaff
  \label{eq:lie-2-var-app}
\end{align}
\begin{align}
  \cov(\dynAff(\bfx)\ctrlaff, \grad \LieOne h(\bfx)^\top \dynAff \ctrlaff)
  &=
    \grad \cov(\dynAff(\bfx)\ctrlaff, \LieOne h(\bfx))  \bfM_0(\bfx)\ctrlaff
    +  \ctrlaff^\top\bfB(\bfx, \bfx)\ctrlaff\bfA  \grad \E[\LieOne h(\bfx)]
  \\
  \cov(\LieOne(\bfx), \grad \LieOne h(\bfx)^\top \dynAff \ctrlaff)
  &=
    \grad \Var(\LieOne(\bfx)) \bfM_0(\bfx)\ctrlaff
    + \cov(\LieOne(\bfx), \dynAff(\bfx)\ctrlaff) \grad \E[\LieOne h(\bfx)]
  \label{eq:lie-2-lie-1-covar-app}
\end{align}%
%

%
%

\begin{algorithm2e}
  \DontPrintSemicolon
  \LinesNumbered
  \SetAlgoLined
  \KwData{Training data $\StDat$, $\CtDat$ at discretization interval $\tau$.
Gaussian process priors $\bfA$ and $\bfB_0(\bfx, \bfx')$. Test state $\bfx_*$ and
$\bfu_*$.}
  \KwResult{$\mCBCtwo$ and $\VCBCtwo$}
  Compute approximate state time derivative
  $\stdt_t \leftarrow \frac{\bfx_{t+1} - \bfx_{t}}{\tau} $ for all $t \in [1,
  \dots, d-1]$. \;
  Collect $\StDtDat = [\stdt_1^\top, \dots, \stdt_{d-1}^\top]^\top$. \;
  Compute $\mDynAffs(\bfx_*)$ and $\bfBs(\bfx_*, \bfx_*)$ from
  \eqref{eq:mvg-posterior}. \;
  Compute mean and variance of $\dynAff(\bfx)\ctrlaff$ from
  \eqref{eq:mvg-posterior-bfu}. \;
  Compute mean and variance of $\LieOne h(\bfx)$ using
  \eqref{eq:mean-var-lie-grad-h-f}
  \;
  Compute mean, variance and covariance of $\grad \LieOne h(\bfx) \dynAff_k \ctrlaff$ using
  \eqref{eq:lie-2-mean-app},
  \eqref{eq:lie-2-var-app} and
  \eqref{eq:lie-2-lie-1-covar-app} respectively.
  \;
  Plug the above values into \eqref{eq:mean-var-cbc} to get $\mCBCtwo$ and $\VCBCtwo$.
  \caption{Algorithm to compute Mean and variance of CBF of relative degree 2}
\label{alg:Algo}
\end{algorithm2e}

%% file: tex/mean-and-variance-of-cbf-2/differentiating-gp.tex
\subsubsection{Differentiating Random Process (Lemma~\ref{lemma:differentiating-gp})}
\label{proof:differentiating-gp}

Let $q(\bfx)$ be a random process with finite $\mu(\bfx)$ mean and finite variance $\knl(\bfx, \bfx')$ for all $\bfx \in \calX$. 
Next we compute the mean and variance of random process $\grad_\bfx q(\bfx)$,
\newcommand{\dbfy}{\delta \bfy}
\newcommand{\dbfx}{\delta \bfx}
\begin{align}
\E[\grad q(\bfx)] 
&= \lim_{\dbfx \to 0} \frac{1}{\dbfx}\E[ q(\bfx + \dbfx) - q(\bfx)]
= \lim_{\dbfx \to 0} \frac{1}{\dbfx}\E[ q(\bfx + \dbfx)] - \E[q(\bfx)]
\\
&= \lim_{\dbfx \to 0} \frac{1}{\dbfx}\mu(\bfx + \dbfx) - \mu(\bfx)
= \grad \mu(\bfx)
\end{align}%

To compute variance, consider a random process with zero mean, $ z(\bfx) = q(\bfx) -
\mu(\bfx)$ but same variance as $q$.
We write differentiation as limit of finite differences,
\begin{align}
  \Var(\grad q) &= \Var(\grad z) = \E[\grad z(\bfx) \grad z(\bfy)^\top]
                  = \lim_{\dbfx,\dbfy \to 0}\E\left[
  \frac{z(\bfx + \dbfx) - z(\bfx)}{\dbfx}
  \left(\frac{z(\bfy + \dbfy) - z(\bfy)}{\dbfy}\right)^\top
  \right],
  \\
  \text{ where } & \frac{z(\bfx + \dbfx) - z(\bfx)}{\dbfx} = \begin{bmatrix}
\frac{z(\bfx + \dbfx) - z(\bfx)}{\delta x_1} & \dots & \frac{z(\bfx + \dbfx) - z(\bfx)}{\delta x_n}.
    \end{bmatrix}
\end{align}%

Considering only $i,j$th element of variance
\newcommand{\dyj}{\delta y_j}
\newcommand{\dxi}{\delta x_i}
\begin{align}
  [\Var(\grad q)]_{i,j}
  &=
    \lim_{\dxi, \dyj \to 0} \E\left[
  \frac{z(\bfx + \dbfx) - z(\bfx)}{\dxi}
  \frac{z(\bfy + \dbfy) - z(\bfy)}{\dyj}
                          \right]
    \\
    &=
    \lim_{\dxi, \dyj \to 0} 
      \frac{1}{\dxi\dyj}\Bigl(\E[z(\bfx + \dbfx)z(\bfy + \dbfy)]
      - \E[z(\bfx)z(\bfy + \dbfy)] 
      \nonumber
  \\
      & \qquad - \E[z(\bfx + \dbfx)z(\bfy)] + \E[z(\bfx)z(\bfy)] \Bigr)
  \\
  &=
    \lim_{\dxi, \dyj \to 0} 
      \frac{1}{\dxi\dyj}\Bigl(\knl(\bfx + \dbfx, \bfy + \dbfy)
    - \knl(\bfx + \dbfx, \bfy)
      - [\knl(\bfx, \bfy + \dbfy) - \knl(\bfx, \bfy)] \Bigr)
  \\
  &=
  \frac{ \p^2 \knl(\bfx,\bfy) }
  {\p x_i \p y_j }
\end{align}%
Hence, the variance is the Hessian of the kernel $\Var(\grad q) = H_{\bfx,
  \bfx'}\knl(\bfx, \bfx')$. $\grad q$ can be written as a Vector-variate Gaussian  Process,
\begin{align}
\grad q &\sim \GP(\grad \mu, \knl_{\grad q}) & \text{where } \knl_{\grad q}(\bfx, \bfx') &= H_{\bfx, \bfx'} \knl(\bfx, \bfx')
\end{align}%

We assume another vector-valued Gaussian process $s(\bfy)$ with mean
$\bar{s}(\bfy)$ and known covariance with $q(\bfx)$ as $\cov_{q,s}(\bfx, \bfy)$.
To prove that $ \cov(\grad q(\bfx), s(\bfy)) = \grad_\bfx\cov_{q,s}(\bfx,
\bfy)$, we start with
\begin{align}
  \cov(\grad q, s)
  &= \E[ (\grad q(\bfx) - \grad \mu(\bfx)) (s(\bfy) - \bar{s}(\bfy))^\top ]
    = \E[\grad q(\bfx) s(\bfy)^\top ] -  \grad \mu(\bfx) \bar{s}(\bfy)^\top
    \label{eq:diff-gp-cov-with-mean}
\end{align}

\begin{align}
  \E[\grad q(\bfx) s(\bfy)^\top ]
  &= \lim_{\dbfx\to 0}\E\left[
  \frac{q(\bfx + \dbfx) - q(\bfx)}{\dbfx} s(\bfy)^\top
  \right],
  \\
  &= \lim_{\dbfx\to 0}
  \frac{\E[q(\bfx + \dbfx)s(\bfy)^\top] - \E[q(\bfx) s(\bfy)^\top]}{\dbfx}
  \\
  &=
    \lim_{\dbfx\to 0} \frac{\cov_{q,s}(\bfx + \dbfx, \bfy) + \mu(\bfx + \dbfx)\bar{s}(\bfy)^\top - \cov_{q,s}(\bfx, \bfy)- \mu(\bfx)\bar{s}(\bfy)^\top }{\dbfx}
  \\
  &=
    \grad_\bfx \cov_{q,s}(\bfx, \bfy) + \grad_\bfx \mu(\bfx)\bar{s}(\bfy)^\top
    \label{eq:diff-gp-cov-no-mean}
\end{align}%
From \eqref{eq:diff-gp-cov-with-mean} and \eqref{eq:diff-gp-cov-no-mean}, we get
$\cov(\grad q(\bfx), s(\bfy)) = \grad_\bfx \cov_{q,s}(\bfx, \bfy)$.

To prove that $\cov(\grad q(\bfx), q(\bfx)) = \grad_\bfx \knl(\bfx, \bfx')$,
substitute $s = q$.
\hfill $\blacksquare $

%% file: tex/mean-and-variance-of-cbf-2/differentiating-lie.tex
\subsubsection{Differentiating Gradient of Lie~\eqref{eq:mean-var-grad-lie-grad-h-f-expanded}}
\label{proof:differentiating-lie}

\begin{align}
    \calH(\bfx, \bfy) := H_{\bfx, \bfy} \left[ \knl(\bfx, \bfy)\grad h(\bfx)^\top K^f \grad h(\bfy) \right]
\end{align}

\newcommand{\pd}[2]{\frac{\partial #1}{\partial #2}}
\newcommand{\pdd}[3]{\frac{\partial #1}{\partial #2 \partial #3}}
\begin{align}
    \left[ \calH(\bfx, \bfy) \right]_{i,j} &= \pdd{}{x_i}{y_j}  \left[ 
    \knl(\bfx, \bfy)\grad h(\bfx)^\top K^f \grad h(\bfy) 
    \right]
    \\
    &= \pd{}{x_i} \left[ 
    \pd{\knl(\bfx, \bfy)}{y_j} \grad h(\bfx)^\top K^f \grad h(\bfy) 
    + \knl(\bfx, \bfy)\grad h(\bfx)^\top K^f \pd{\grad h(\bfy)}{y_j}
    \right]
    \\
    &= 
    \pdd{\knl(\bfx, \bfy)}{x_i}{y_j} \grad h(\bfx)^\top K^f \grad h(\bfy) 
    + \pd{\knl(\bfx, \bfy)}{y_j} \pd{\grad h(\bfx)^\top}{x_i} K^f \grad h(\bfy) 
    \\
    &\qquad
    + \pd{\knl(\bfx, \bfy)}{x_i}\grad h(\bfx)^\top K^f \pd{\grad h(\bfy)}{y_j}
    + \knl(\bfx, \bfy)\pd{\grad h(\bfx)^\top}{x_i} K^f \pd{\grad h(\bfy)}{y_j}
\end{align}

Considering matrix forms of each of the terms in above equation, 
\begin{align}
[H_{\bfx, \bfy}\knl(\bfx, \bfy)]_{i,j} \grad h(\bfx)^\top K^f \grad h(\bfy) ]_{i,j}
&= 
    \pdd{\knl(\bfx, \bfy)}{x_i}{y_j} \grad h(\bfx)^\top K^f \grad h(\bfy) 
    \\
    [ \grad h(\bfy)K^{f\top}  H_{xx} h(\bfx) \grad_\bfy \knl(\bfx, \bfy)^\top]_{i,j} 
    &=  \pd{\knl(\bfx, \bfy)}{y_j} \pd{\grad h(\bfx)^\top}{x_i} K^f \grad h(\bfy) 
    \\
    [\grad_\bfx \knl(\bfx, \bfy) \grad h(\bfx)^\top K^f H_{yy} h(\bfy)]_{i,j} 
    &= \pd{\knl(\bfx, \bfy)}{x_i}\grad h(\bfx)^\top K^f \pd{\grad h(\bfy)}{y_j}
    \\
    [\knl(\bfx, \bfy) H_{xx} h(\bfx) K^f H_{yy} h(\bfy)]_{i,j}
    &= \knl(\bfx, \bfy)\pd{\grad h(\bfx)^\top}{x_i} K^f \pd{\grad h(\bfy)}{y_j}
\end{align}

Hence,
\begin{align}
\calH(\bfx, \bfy) 
&= H_{\bfx, \bfy}\knl(\bfx, \bfy)]_{i,j} \grad h(\bfx)^\top K^f \grad h(\bfy)
+ \grad h(\bfy)K^{f\top}  H_{xx} h(\bfx) \grad_\bfy \knl(\bfx, \bfy)^\top
\\
&\qquad
+ \grad_\bfx \knl(\bfx, \bfy) \grad h(\bfx)^\top K^f H_{yy} h(\bfy)
+ \knl(\bfx, \bfy) H_{xx} h(\bfx) K^f H_{yy} h(\bfy)
\end{align}